%% file: main.tex
\documentclass{article}

\usepackage[preprint, nonatbib]{neurips_2023}

\usepackage[utf8]{inputenc} 
\usepackage[T1]{fontenc}    
\usepackage{hyperref}       
\usepackage{url}            
\usepackage{booktabs}       
\usepackage{amsfonts}       
\usepackage{amsmath}
\usepackage{nicefrac}       
\usepackage{microtype}      
\usepackage{xcolor}         
\usepackage{amssymb} 

\usepackage{graphicx}
\usepackage{amsmath}
\usepackage{bbm}
\usepackage{algorithm}
\usepackage{algorithmic}
\usepackage{caption}
\usepackage{subcaption}

\usepackage{bibentry}
\newtheorem{theorem}{Theorem}[section]
\newtheorem{corollary}{Corollary}[theorem]

\newtheorem{remark}[corollary]{Remark}

\DeclareMathOperator*{\argmin}{arg\,min}

\title{Extreme Risk Mitigation in Reinforcement Learning using Extreme Value Theory}

\author{%
 Karthik Somayaji NS\\
  University of California, Santa Barbara, USA\\
  \And
  Yu Wang\\
  University of California, Santa Barbara, USA\\
  \And
  Malachi Schram \\
  Jefferson National Laboratory, VA, USA \\
  \And
  J\'an Drgo\v na \\
  Pacific Northwest National Laboratory, WA, USA\\
  \And
  Mahantesh Halappanavar \\
  Pacific Northwest National Laboratory, WA, USA \\
  \And
  Frank Liu \\
  Oak Ridge National Laboratory, TN, USA\\
  \And
  Peng Li \\
  University of California, Santa Barbara, USA\\
  }

\begin{document}

\maketitle

\begin{abstract}
Risk-sensitive reinforcement learning (RL) has garnered significant attention in recent years due to the growing interest in deploying RL agents in real-world scenarios. A critical aspect of risk awareness involves modeling highly rare risk events (rewards) that could potentially lead to catastrophic outcomes. These infrequent occurrences present a formidable challenge for data-driven methods aiming to capture such risky events accurately. While risk-aware RL techniques do exist, their level of risk aversion heavily relies on the precision of the state-action value function estimation when modeling these rare occurrences. Our work proposes to enhance the resilience of RL agents when faced with very rare and risky events by focusing on refining the predictions of the extreme values predicted by the state-action value function distribution. To achieve this, we formulate the extreme values of the state-action value function distribution as parameterized distributions, drawing inspiration from the principles of extreme value theory (EVT). This approach effectively addresses the issue of infrequent occurrence by leveraging EVT-based parameterization. Importantly, we theoretically demonstrate the advantages of employing these parameterized distributions in contrast to other risk-averse algorithms. Our evaluations show that the proposed method outperforms other risk averse RL algorithms on a diverse range of benchmark tasks, each encompassing distinct risk scenarios.
\end{abstract}

\section{Introduction}

In the recent years, there has been a wide array of research in leveraging reinforcement learning as a tool for enabling agents to learn desired behaviors. Learning such specific behaviors may entail satisfying some very specific constraint sets \cite{SRL_CMDP_RL} in the underling Markov decision process (MDP) or may include identification of safe and unsafe regions of operation \cite{SRL_LeaveNT}. The learnt behavior may be trained even to be robust to noisy environments with injected perturbations \cite{SRL_RobustAR, SRL_DistributionallyRR, SRL_D_robust_MDP}. However, a central theme in most recent works in reinforcement learning has been in optimizing the future returns while ensuring that the behavioral policy satisfies some constraints. A natural consequence of having to tune the agent's behavior lies in training the agent to be either risk seeking or risk averse. Risky states or returns are naturally not preferred in most safety critical applications, for instance in self driving, robotic surgery, particle accelerator controllers etc. Thus risk quantification in the context of the learning agent becomes crucially important in avoiding catastrophic failures. 

Although numerous risk quantifying RL algorithms exercise precise control by mitigating risk, they are very susceptible to ignoring risk in the low data regime as suggested by \cite{EVT2_CI, EVT1_beranger}. In the eventuality of scarce data availability, the estimates of risk tend to be high variance estimates of the true risk and thus tend to be unreliable for any optimization. Many risk averse RL techniques quantify extreme risks based on the estimation of the extreme value quantiles of the state action value function. For low risk events, the distribution of returns for instance, can be arbitrarily skewed or heavy tailed. The natural scarcity of data in this scenario casts uncertainty in the estimate of the risk measure. 

Thus, there is an acute need to improve the prediction of extreme values of the state action value function distribution. The inherent scarcity of data in rare events makes the problem of estimating the quantiles of the state action value distribution non-trivial. In this work, we propose to develop methods that help in extreme risk mitigation by modelling the state action value function appropriately.  In particular, we develop an extreme value theory (EVT) based method to model the tail region of the quantile of the state action value function. Extreme value theory (EVT) posits that the  tail distribution obeys a certain parametric form, which we propose to fit for the particular case of the state action value distribution. We theoretically and empirically prove the effectiveness of our proposed method. Thus, our contributions include :
\begin{enumerate}
    \item Identifying the drawbacks of using the popular quantile regression frameworks like\cite{RA_DSAC_risk, RA_quantile_RL, RA_raac} to model the state action value functions which are likely to have rare extreme values.  
    
    \item Modelling the tail region of the state action  value function as a parameterized distribution drawing inspiration from extreme value theory (EVT) using a (General Pareto distribution) so as to better model the extreme realizations even in the low data regime. 

    \item Proving reduction in variance of the quantile estimators of the state action value distribution when employing  extreme value theory based methods.
    
\end{enumerate}

\section{Related Work}
There has been an extensive study in incorporating reinforcement learning in risk avoidance. \cite{RM_CVAR_actor_critic, RM_policy_grad_cvar, RM_risk_sensitive_robust_decision} have studied quantification of risk in terms of percentile criteria, specifically the CVaR (conditional value at risk). Other risk averse measures also include \cite{RM_variance1, RM_variance2} which incorporate variance as a risk measure, while \cite{RM_ConstrainedRL} use a notion of range to discuss risk. All the above metrics do require an understanding of the distribution of the quantity of interest.  Most works in the RL context, deal with risk as it is connected to the distribution of the state action value function (or the Q-function). The work of \cite{RA_DRL} offers a distributional perspective on the value functions. \cite{RA_quantile_RL} approximate the distribution of the value functions in terms of the quantiles of the distribution. Other works like \cite{RA_DSAC_risk, RA_raac, RA_wcpg} use the notion of quantiles of the value function distribution to quantify risk.
 However, \cite{RA_safety_critic} also discuss methods for fine tuning agents in transfer learning scenarios. Although many of these works perform well with sufficient data, much attention has not been paid to risk aversion under extremely rare catastrophic risks.

 When addressing rare risky events, extreme value theory (EVT) \cite{EVT_text} provides a framework to characterizing the asymptotic behavior of the distribution of the sample extrema. EVT finds extensive application in modeling rare events across domains from finance, to operations research and  meteorology to multi-armed bandit problems \cite{EVT_text2, EVT_appl1, EVT_appl2, EVT3_bandits}.

 In the reinforcement learning context, \cite{EQL} recently discuss the use of EVT in estimating the maximum of the Q-value in the Bellman update.  Our work is different from \cite{EQL} on three main fronts. (i) Firstly, on the applications side, \cite{EQL} aim to learn the optimal value function in the maximum entropy RL setting, inspired by principles in EVT. However, our work aims to perform risk averse decision making in extremely rare risky scenarios. (ii)\cite{EQL} use the principles from EVT, particularly the fact that the sample maxima can be characterized by the Gumbell distribution to estimate the max entropy optimal value function. However, our work considers modelling the entire state action value distribution precisely by using the principle of asymptotic conditional excess distributions to estimate the underlying tail behavior of the distribution. (iii) \cite{EQL} use Gumbel regression as a tool to define and train a soft Q-function by gathering inspiration from the Gumbell distributed Bellman errors. This is accomplished by using the Fisher Tippet Theorem (Theorem \ref{th:FTT}) which provides the limiting behavior of sample maxima. In our work, we estimate the realizations of the critic (state action value) distribution over a threshold by employing another key theorem in extreme value theory namely Theorem \ref{th:PBD}. Particularly, our method uses maximum likelihood estimation of the GPD distribution to estimate the state action value distribution.

 Thus, our work although using EVT, is different in application, methodology and theoretical reasoning to \cite{EQL}.

 \input{method}

\input{supplementary}

\bibliography{main}

\end{document}

%% file: method.tex
\section{Notation}
We introduce the notation for the standard Markov Decision Process (MDP) characterized by the tuple $(\mathcal{S}, \mathcal{A}, P_R, P_S, \gamma)$, where $\mathcal{S}$ is the state space, $\mathcal{A}$ is the action space, $P_R$ is the stochastic reward kernel such that $P_R: \mathcal{S} \times \mathcal{A} \to \mathcal{P}(\mathcal{R})$, where $\mathcal{R}$ is the reward set. $P_S: \mathcal{S} \times \mathcal{A} \to \mathcal{P}(\mathcal{S}) $ is the probabilistic next state transition kernel, and $\gamma$ is the discount factor. The policy $\pi$ of the agent is a mapping $\pi: \mathcal{S} \to \mathcal{P}(\mathcal{A})$. The future sum of discounted returns is a random variable denoted by $J^{\pi}(s,a) = \sum_{t=0}^{\infty}\gamma^tR_t$, where $R_t \sim P_R(S_t, A_t)$ and $A_t \sim \pi(S_t)$ with $S_0=s; A_0=a$.We denote the distribution corresponding to the random variable $J^{\pi}$ as $Z^{\pi}$. The terms $S_t, A_t, R_t$ are the encountered state, action and reward, respectively at time step $t$.

\section{Background}

We provide a brief background on bothe distributional reinforcement learing and extreme value theory in this section

\subsection{Distributional Reinforcement Learning}

Reinforcement learning involves tuning the behavior of the agent so as to achieve maximum cumulative reward. The state action value function distribution $Z$ under policy $\pi$ is updated using the distributional Bellman operator
\begin{equation}
\label{eqn:Bellman_op}
    T^{\pi}Z(s,a) = r(s,a) + \gamma \mathrm{E}_{s' \sim P_S(s,a),a' \sim \pi(s')}Z(s',a')
\end{equation}

The Bellman operator $T^{\pi}: \mathcal{P}(\mathrm{R}^{\mathcal{S} \times \mathcal{A}}) \to \mathcal{P}(\mathrm{R}^{\mathcal{S} \times \mathcal{A}})$ operates on the space of probabilities over the reals $\mathrm{R}$, for each state action pair. The update of $Z^{\pi}(s,a)$ can be viewed as a scale and shift operation of the mixture distribution $\mathrm{E}_{s' \sim P_S(s,a),a' \sim \pi(s')}Z(s',a')$. 

The distribution function $Z^{\pi}(s,a)$ characterizes the values that the random variable $J^{\pi}(s,a)$ can assume. Thus, a knowledge of the distribution function $Z^{\pi}(s,a)$ aids in understanding even the extreme values that $J^{\pi}(s,a)$ can be assigned. Focusing attention on Eqn.\ref{eqn:Bellman_op} reveals that the update of the LHS distribution $Z^{\pi}(s,a)$ happens via sampling the mixture distribution $\mathrm{E}_{s' \sim P_S(s,a), a' \sim \pi(s')}Z(s',a')$, scaling it by $\gamma$ and shifting it by $r(s,a)$, which is the sampled scalar reward. 

\subsection{Extreme Value Theory (EVT)}
\label{section:EVT}
As discussed earlier, EVT is a tool for characterizing the extreme behavior of distributions. For the purpose of brevity, we introduce the two main theorems in EVT.

\begin{theorem}[Fisher Tippet Theorem \cite{EVT_FTT}]
\label{th:FTT}
 Let $X_1, \cdots ,X_n$ be a sequence of IID random variables, with a distribution function (CDF) $F$. Let $M_n$ represent the sample maximum. If there exist constants $a_n > 0$ and $b_n$ and a non-degenerate distribution function $G$ such that:
\[\lim_{n \to \infty} P\Big\{ \frac{M_n - b_n}{a_n} \leq x  \Big\} = G(x), \]
then the distribution function $G(x)$ is called the Generalized Extreme Value distribution (GEV) and can be decomposed into either the Gumbell, Frechet or the Weibull distribution.
\end{theorem}

 Intuitively the Fisher Tippet Theorem describes that the normalized sample maxima of a distribution $F$, converges in distribution to the GEV distribution. 

The other central theorem in EVT is the  Pickands-Balkema -de Haan theorem (PBD theorem) which inspects the conditional exceedance probability above a threshold $u$, of a random variable $X$ with a distribution function $F$.

\begin{theorem}[Pickands-Balkema-de Haan Theorem ]
\label{th:PBD}
\cite{EVT_PBD} Let $X_1 \cdots X_n$ be a sequence of IID random variables with distribution function (CDF) given by $F$ whose limiting behavior approaches the GEV distribution. Let $F_u(x) = P(X-u \leq x| X>u)$ be the conditional excess distribution. Then,
\[\lim_{u \to \infty}F_u(x) \xrightarrow[]{D} H_{\xi, \sigma}(x),\]
where $H_{\xi, \sigma}(x)$ is the Generalized Pareto distribution (GPD) with parameters $\xi, \sigma$.
\end{theorem}

Intuitively Theorem\ref{th:PBD} describes that the conditional distribution $F_u$ of the random variable $X$ approaches the GPD distribution for large enough $u$. 
The CDF of the GPD distribution $F_{\xi, \sigma}(x)$ is given by:
\begin{equation}
    \begin{cases}
  1- \Big( 1+ \frac{\xi x}{\sigma} \Big)^{-1/\xi} \; \mathrm{for} \; \xi \neq 0\\    
  1- \exp(-\frac{x}{\sigma}) \; \mathrm{for} \;\xi =0    
\end{cases}
\end{equation}

\section{Motivating Example}

\begin{figure}[!htbp]
\vspace{-0.25cm}
\begin{center}
\includegraphics[width=7cm, height=5cm]{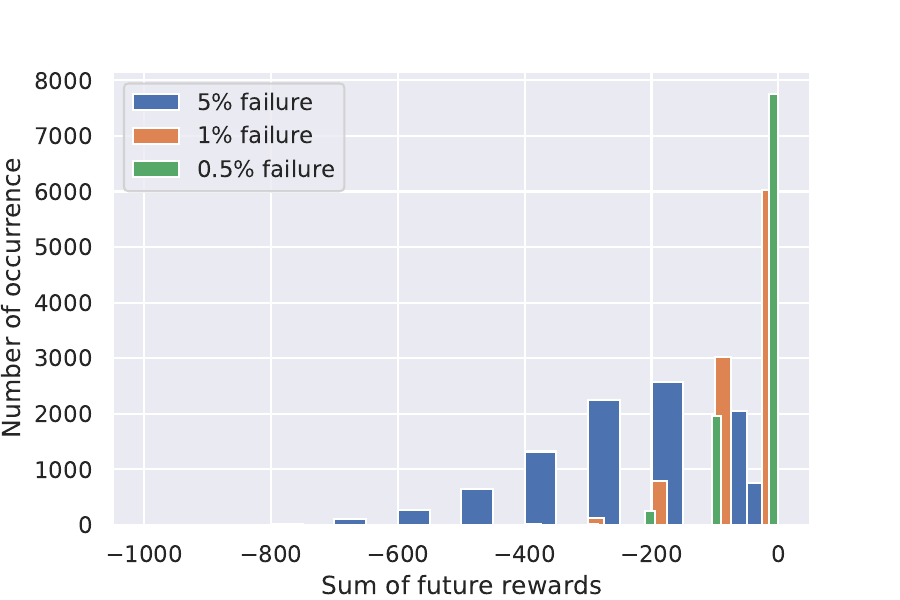}
\end{center}
\caption{The effect of failure frequency $p$ on the sum $J_p$ }
\label{fig:failure_analysis_toy}
\vspace{-0.25cm}
\end{figure}

Many real world scenarios are marked by rare failures, which implies that an inherently low probability mass is assigned to the occurrence of such low reward (failure) cases. The sum of future discounted rewards is consequently characterised by failure events with low probability. A real world example is of a car speeding on a highway. Higher speeds produce higher rewards (in terms of reaching the destination faster), but may rarely lead to catastrophes due to slippery roads for instance. Although the occurrence of such events is sporadic, the outcomes may be very unfavourable. We provide a simple illustration in Figure. \ref{fig:failure_analysis_toy}, where we define $J_p = \sum_{i=1}^{H}a\cdot (s_i \cdot L)$, where $s_i \sim \mathrm{Ber}(p)$, $H=50$, $a=10$ is a constant reward, $L=-10$ is the strength of the penalty and $\mathrm{Ber}(p)$ denotes the Bernoulli random variable with parameter $p$ (which denotes the failure frequency). We plot a histogram of the sum $J_p$ for different $p=0.5\%, 1\%, 5\%$ by simulating the sum 10,000 times. \textbf{It is evident that as $p$ decreases, the $J_p$ values accumulate towards the mode while having very less probability mass in the extreme regions.}

Thus a sampling based approach to updating Eqn.\ref{eqn:Bellman_op}, requires many more samples to accurately represent the true distribution in case of rare failures (low rewards). The low data availability in extreme risk reward scenarios  makes it challenging to update the state action value function. The so modelled state action value distribution function $Z^{\pi}(s,a)$ poorly models the extreme regions of the true underlying distribution under limited data. We recognize that under the limited availability of data, it is non-trivial to model the extreme values of the state action value function.

\section{Drawbacks of Quantile Regression}
\label{sec:QR_drawback}
One of the popular methods to model the distribution of the state action value function is the quantile regression framework used by \cite{RA_DSAC_risk, RA_quantile_RL, RA_raac}. Quantile regression proposed by \cite{QR_OG} estimates the true quantile value of a distribution by minimizing the pinball loss. Assume a random variable $Y$ with its unknown true distribution function $F_Y(.)$ and probability density function $f_Y(.)$. Our interest lies in estimating the true quantile values of $F_Y$ denoted by $\theta_{\tau}$ for $\tau \in [0,1]$, the quantile level. The quantile predicted by the quantile regression framework is a unique minimizer of the pinball loss $\mathcal{L}(\theta_{\tau})$ given by
\begin{align*}
    \mathcal{L}(\theta_{\tau}) = \mathrm{E}_{y \sim F_Y}(y-\theta_{\tau}) (\tau - \mathbbm{1}_{y-\theta_{\tau} < 0})
\end{align*}

Assume $N$ samples sampled from $F_Y$, $\{y_1, y_2, \cdots , y_N\}$. The aim is to find the quantile value estimate $\theta_{\tau}^N$ using quantile regression, by minimizing the empirical pinball loss for a given quantile level $\tau$:

\begin{equation}
 \label{eqn:emp_pinball_loss}
    \mathcal{L}(\theta^{\tau}_N) = \frac{1}{N} \sum_{i=1}^{N} (y_i-\theta_{\tau}^N) (\tau - \mathbbm{1}_{y_i-\theta_{\tau}^N < 0}).
\end{equation}

In deep reinforcement learning, a modified smooth loss function called the empirical quantile Huber loss \cite{RA_huber_loss} is instead used for better gradient back propagation. Importantly, the asymptotic convergence of the quantile regression estimator to the true quantile value is discussed in \cite{QR_OG} as 
\[\sqrt{N} (\theta_{\tau}^N - \theta_{\tau} ) \to^{D} \mathcal {N}(0, \lambda^2),\] where 
the variance of the estimator $\theta^N_{\tau}$ is given by 
\begin{equation}\label{eqn_qr_variance}
\lambda^2 = \frac{\tau(1-\tau)}{f_Y^2(\theta_{\tau})}.
\end{equation}

$\theta^N_{\tau}$ is dependent on the quantile level $\tau$. For $\tau \to 1$ and $\tau \to 0$, the variance reduces. However, the variance is also inversely dependent on the squared probability density at the quantile value $\theta_{\tau}$. Thus for the same quantile level $\tau \to 1^-$, for instance, the variance of the estimator depends on the probability density at the high quantile levels. Thus, in case of rare extreme valued state action value functions, the quantile regression estimate of the true quantile has high variance. \cite{EVT1_beranger, EVT2_CI} also discuss the estimation inaccuracy of the quantile estimates under lower data availability. \cite{EVT6_undercoverage} specifically discusses the inherent undercoverage associated with quantile regression estimator for high quantile regions. Such evidence coupled with the high variance property acts as a deterrent to exclusively  choosing quantile regression in estimating extreme quantiles in the case of distributions which also model low probability extreme events. 

To this end we propose the first extreme value theory based actor-critic distributional reinforcement learning for extreme risk mitigation. 

\section{Extreme Value Theory in Distributional Reinforcement Learning}

Extremely rare events tend to make the state action value distribution be characterized by low probability tails (extreme events). The higher probability occurrences of the true random return $J^{\pi}(s,a)$ can be learnt reasonably well, unlike the low probability extreme valued realizations. Although the non-tail region can be learnt through data sampling, the tail region needs special handling under limited data availability. 

\begin{remark}
\label{rmk: negate}
Although higher rewards are considered better in reinforcement learning, we negate the rewards to maintain consistent notation with literature in EVT. This does not affect any analysis. Because of the negated reward, we swap focus from the left tail of the state action value distribution to the right tail.    
\end{remark}

To address the discussed challenge, we propose to invoke the Pickands-Balkema-de Haan theorem (Theorem.\ref{th:PBD}) to approximate the tail regions of the state action value function distribution $Z^{\pi}$. The advantage of using such an approximation is that the tail region (extreme value) behavior of $Z^{\pi}$ can be captured even under lower data availability. Thus, approximating the tail region of $Z^{\pi}$ as the conditional excess has the overall effect of better representing the true state action value distribution unlike the conventional data sampling techniques. Another advantage of using the conditional excess distribution is the power to extrapolate to unseen data points on the support of $Z^{\pi}$. The conditional excess as stated in Theorem \ref{th:PBD}, can be represented by the GPD distribution with parameters $\xi, \sigma$. Thus, for the state action value distribution $Z^{\pi}(s,a) , \forall (s,a) \in \mathcal{S} \times \mathcal{A}$, the tail region can be modelled by a corresponding GPD distribution with parameters $\xi(s,a), \sigma(s,a) ,\forall (s,a) \in \mathcal{S} \times \mathcal{A}$. Thereby, the tail regions of each $Z^{\pi}(s,a)$ can equivalently be viewed as paramterized GPD distributions in $\mathcal{S} \times \mathcal{A}$ space, i.e., $\mathrm{GPD}\Big(\xi(s,a), \sigma(s,a) \Big)$ with different thresholds $u$ for each $(s,a)$. 

\section{Extreme Valued Actor - Critic (EVAC) }

\begin{figure*}[!htbp]
\begin{center}
\includegraphics[width=10cm, height=4.5cm]{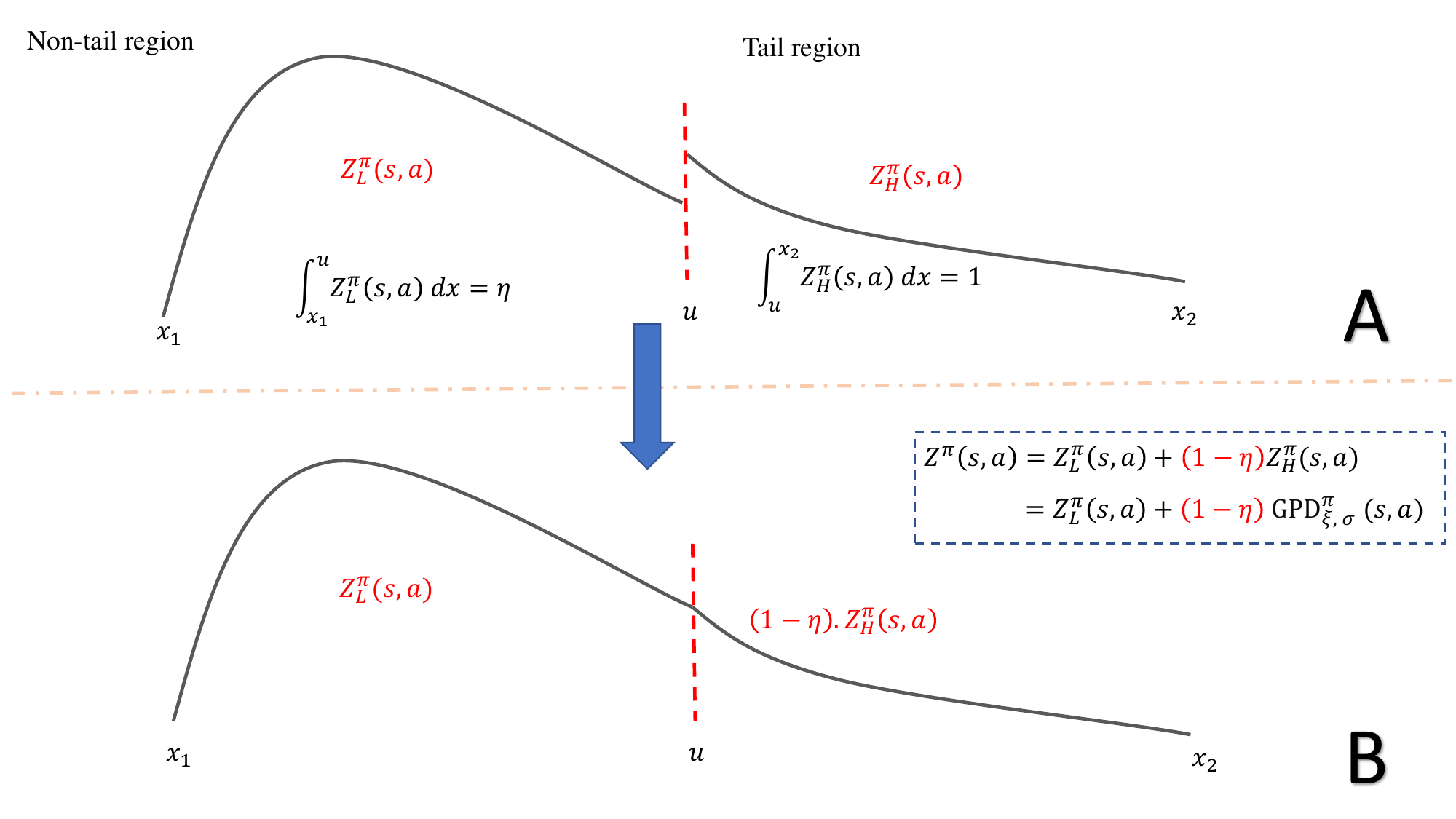}
\end{center}
\caption{Modeling the tail and non-tail distributions of the state action value function. The area under the non-tail distribution $Z^{\pi}_L(s,a)$ is $\eta$. The area under the tail region of the distribution $Z^{\pi}_H(s,a)$ is 1. }
\label{fig:tail_non_tail}
\end{figure*}

As shown in Fig.~\ref{fig:tail_non_tail}, for a given threshold $u$, we denote by $Z^{\pi}_L(s,a)$, the non-tail distribution. The subscript '$L$' is used for denoting support values of $Z^{\pi}$ lower than $u$. Similarly, we denote by $Z^{\pi}_H(s,a)$, the tail distribution. The subscript '$H$' denotes distribution with support values higher than $u$. $u$ is assumed to be a sufficiently high threshold for the state action pair $(s,a)$. 
 We assume the area under $Z^{\pi}_L(s,a)$ to be $\eta$. The area under $Z^{\pi}_H(s,a)$ is assumed to be 1. So, the state action value distribution $Z^{\pi}(s,a)$ is obtained by rescaling $Z^{\pi}_H(s,a)$.
\begin{equation}
\label{eqn:normalized_Z}
Z^{\pi}(s,a) = Z^{\pi}_L(s,a) + (1-\eta) Z^{\pi}_H(s,a)
\end{equation}
Thus the tail distribution $Z^{\pi}_H(s,a)$ can be modelled by $\mathrm{GPD}\Big(\xi(s,a), \sigma(s,a) \Big)$ in the limit and can be suitably rescaled to obtain $Z^{\pi}(s,a)$ as in Eqn.\ref{eqn:normalized_Z}.

\subsection{Training the state action value function}
In general the Bellman update can be written as:
\begin{equation}
\label{eqn:Bellman_op_EVT}
    T^{\pi}Z(s,a) = r(s,a) + \gamma \mathrm{E}_{s',a'} \Big[ Z_L(s',a') + (1-\eta) Z_H(s',a') \Big]
\end{equation}

The EVT based state action value distribution can be trained by regular quantile regression with some modifications. We sample the RHS of Eqn.\ref{eqn:Bellman_op_EVT}, and then update the state action value function $Z(s,a)$  using quantile regression based on the obtained samples.

To understand the empirical training process, we first explain the sampling procedure from quantile estimates of the state action value distribution $Z^{\pi}(s,a)$. We denote the $\tau^{th}$ quantile of $Z^{\pi}(s,a)$ as $Z^{\pi}(s,a)|_{\tau}$. To sample from the distribution $Z^{\pi}(s,a)$ we employ inverse transform sampling where $\tau \sim U(0,1)$. The resulting values of $Z^{\pi}(s,a)|_{\tau =0 \to 1}$ are then equivalent to sampling from the distribution of $Z^{\pi}(s,a)$. 

$Z^{\pi}_L(s',a') = Pr \Big(Z^{\pi}(s',a')) \leq Z^{\pi}(s',a')_{\eta} \Big)$ is a new distribution that can be obtained by sampling $\tau \sim U(0,\eta)$ and querying $Z^{\pi}(s,a)|_{\tau =0 \to \eta}$. The resulting samples are equivalent to sampling from $Z^{\pi}_L(s',a')$. To obtain samples from $\Big[ Z^{\pi}_L(s',a') + (1-\eta) Z^{\pi}_H(s',a') \Big]$, we sample:
\begin{enumerate}
    \item from $Z^{\pi}(s',a')|_{\tau=0 \to \eta}  $ with proportion $\eta$
    \item from $\mathrm{GPD}_{\xi, \sigma}(s',a')$ and shift it by $Z^{\pi}(s',a')|_{\tau=\eta}$ with proportion $1-\eta$. 
\end{enumerate}

The shifted GPD distribution models the tail region of the state action value function while samples from $Z^{\pi}_L(s',a')$ model the non-tail region. Having obtained samples from the RHS of Eqn.\ref{eqn:Bellman_op_EVT}, the state action value function $Z^{\pi}(s,a)$ can be trained using quantile regression. It is to be noted that although the state action value function is modelled by quantile regression, the sampling procedure involves sampling of $Z^{\pi}_L(s'a')$ and the shifted GPD distribution which better models the tail behavior, unlike \cite{RA_DSAC_risk,RA_quantile_RL,RA_raac}. \textbf{We provide a proof of convergence of the Bellman operator in Eqn.\ref{eqn:Bellman_op_EVT} in} \textcolor{red}{Section A of the Appendix}.

However, it is also essential to fit the $\xi(s'a'), \sigma(s',a')$ to the GPD distribution so that Eqn.\ref{eqn:Bellman_op_EVT} is accurately modelled. To train $\xi(s'a'), \sigma(s',a')$, we obtain samples from the tail of the distribution $Z^{\pi}(s',a')$. To obtain excess samples from the state action value distribution, we first obtain samples $Z^{\pi}(s',a')|_{\tau = \eta \to 1}$ and subtract these by  $Z^{\pi}(s',a')|_{\tau = \eta}$ to obtain the excess over the threshold. Through this procedure, we can fit $\xi(s'a'), \sigma(s',a')$ by maximum likelihood estimation of the GPD distribution.
\textbf{We provide a detailed explanation of the MLE procedure of $\xi(s',a'), \sigma(s',a')$ in} \textcolor{red}{Section C of the Appendix}.
Thus, we provide a novel framework for using extreme value theory for state action value distribution estimation.

\subsection{Policy Optimization}
In order to train risk averse policies, extreme values of the state action value function are employed to guarantee optimal worst case performance. We propose to employ the CVaR approach \cite{RM_policy_grad_cvar, SRL_cond_CVAR} on the state action value function $Z^{\pi}(s,a)$ for mitigating extreme risk. The CVaR (Conditional value at risk), is a risk measure that denotes the average worst case performance by integrating the quantiles of the state action value distribution between two quantile levels $x_1$ and $x_2=1.0$. The optimal policy $\pi$ is as in Eqn.\ref{eqn:policy_imp}. Typically we choose $x_1 \gg 0.5$. ( Note the negation of the reward (Remark \ref{rmk: negate}) which paves the way for $x_1 \gg 0.5$). We set $x_2=1.0$.
\begin{align}
\label{eqn:policy_imp}
    \pi^{*} &=  \argmin_{\pi} \mathrm{CVaR(x_1)} \\ \notag
    &=\argmin_{\pi} \frac{1}{x_2 - x_1}\int_{\tau=x_1}^{x_2=1} Z\Big(s,\pi(s)\Big) \Big|_{\tau}
\end{align}

 \textbf{We illustrate the entire algorithmic flow of the EVAC method above in} \textcolor{red}{Section B of the Appendix}. 

\section{Variance reduction in the quantile regression estimate using EVAC}
We propose to model the state action value distribution using quantiles updated through quantile regression. However, as stated earlier in Section \ref{sec:QR_drawback}, the quantile regression estimator's variance is inversely proportional to the square of the probability density at a particular quantile level. A natural question arises about the variance in the quantile regression estimate when the tail region of the underlying distribution is modelled using a GPD approximation. To address this question, we analyse the variance of the quantile regression estimator under the GPD tail and otherwise. 

Assume a random variable $Y$ with its distribution function $F_Y$. Let the $\tau^{th}$ quantile of $F_Y$ be denoted as $\theta_{\tau}$. For any sufficiently large quantile level $\eta \in [0,1)$, and a smaller increment quantile level $t \in [0,1)$ such that, $\eta +t \in [0,1)$, we have the corresponding quantiles of the distribution $\theta_{\eta}$ and $\theta_{\eta + t}$. We define the excess value of the quantile as $x = \theta_{\eta + t} - \theta_{\eta}$.
\begin{align}
 F_Y(\theta_{\eta + t}) &= P\Big(Y \leq \theta_{\eta} +x \Big) \\ \notag
 &= \eta + (1-\eta) P\Big(Y-\theta_{\eta} \leq x | Y > \theta_{\eta} \Big) \\ \notag
 &= \eta + (1-\eta) H_{\xi, \sigma}(x)
\end{align}
As mentioned earlier for sufficiently large $\eta$, $P\big(Y-\theta_{\eta} \leq x | Y > \theta_{\eta} \big) $
 approaches the GPD distribution $H_{\xi, \sigma}(x)$. Thus, we have
 \begin{equation}
 \label{eqn:cdf_of_Y}
     P(Y \leq \theta_{\eta} + x) = \eta + (1-\eta)P(X \leq x)
 \end{equation}
 where $X \sim H_{\xi, \sigma}$.
It follows from Eqn.\ref{eqn:cdf_of_Y}, that,
\begin{equation}
    f_Y (\theta_{\eta} + x) = (1-\eta) f_{H_{\xi,\sigma}}(x)
\end{equation}

If we represent the quantiles of $H_{\xi, \sigma}$ by $\theta^H$, then we have the following relationship between the quantiles of $F_Y$ and the quantiles of the GPD distribution :

\begin{equation}
\label{eqn:linear_theta_relationship}
    \theta_{\eta + t} = \theta{\eta} + \theta^H_{\frac{t}{1-\eta}}
\end{equation}

We are interested in representing the quantiles for sufficiently large $\eta$ and higher. Following Eqn.~\ref{eqn_qr_variance}, the variance of the quantile regression estimator in estimating the quantile $\theta_{\eta + t}$ of the distribution function $Y$ is
\[ \lambda_Y^2 = \frac{(\eta + t)(1-\eta -t)}{(1-\eta)^2 f^2_{H_{\xi, \sigma}} \Big(\theta^H_{\frac{t}{1-\eta}} \Big)} \]

However, from Eqn.\ref{eqn:linear_theta_relationship}, if $\theta_{\eta}$ is known, one may simply estimate the $\frac{t}{1-\eta}^{th}$ quantile of the GPD distribution and shift it by $\theta_{\eta}$. Thus the quantile regression estimator's variance in estimating $\eta +t$ quantile of $Y$ using the GPD is given by:

\[ \lambda_H^2 = \frac{(t/(1-\eta))(1-t/(1-\eta))}{f^2_{H_{\xi, \sigma}} \Big(\theta^H_{\frac{t}{1-\eta}} \Big)} \]

\begin{figure}[!htbp]
\vspace{-0.75cm}
\begin{center}
\includegraphics[width=9cm, height=6cm]{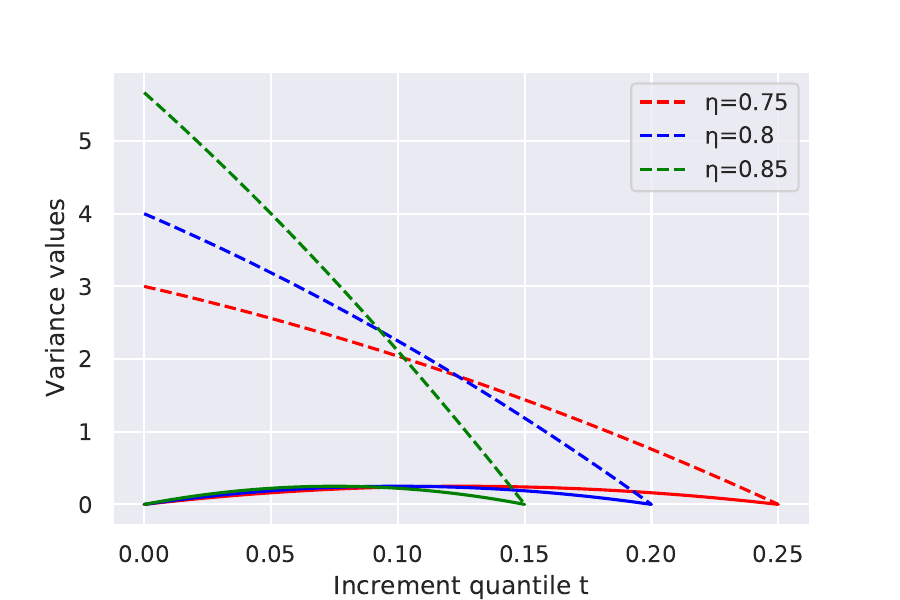}
\end{center}
\caption{Dashed lines indicate $\lambda^2_Y$ and the solid lines indicate $\lambda^2_H$. $\lambda^2_H$ is always upper bounded by the $\lambda^2_Y$ which illustrates variance reduction while using GPD for approximating the tail. }
\vspace{-0.5cm}
\label{fig:variance}
\end{figure}

We can verify that $\lambda_Y^2 \gg \lambda_H^2$ for large values of $\eta$, e.g., close to $1.0$.  
Therefore, we show that GPD based modelling of the tail region of the distribution $Y$ helps reduce the variance in the estimation of higher quantiles. We also illustrate this in Figure.\ref{fig:variance} with a few candidate values for $\eta= 0.75,0.8, 0.85$.

\section{Experimental Evaluation}
In this section, we empirically verify the effectiveness of using EVT in practical reinforcement learning scenarios. We experiment on two benchmark Open-AI environments \cite{gym} namely Mujoco environments and Safety-gym environments \cite{safety-gym}. For creating rare risky events, we modify the reward using a wrapper function which penalizes the reward for certain state action pairs with a certain probability (which is typically small to simulate rare risky events). 
Denoting $R(s,a)$ as the original non-penalized reward, we define the stochastic penalized reward $r(s,a)$ as

\begin{equation*}
\label{eqn:reward_penalty}
    \mathbf{r(s,a) = R(s,a) +} \mathbf{\mathbb{I}_{q > \alpha} L.\mathcal{B}_p}
\end{equation*}

where $\mathbb{I}$ is the indicator function. 
We define the penalty weight term $L$ and a Bernoulli distribution parameter $p$ that indicates the frequency at which the state action pair $(s,a)$ is penalized. The Bernoulli random variable is denoted by $\mathcal{B}_p$.We also set a threshold $\alpha$ which acts as the value above or below which rewards are penalized. Our reward penalization setup is similar to the set up of \cite{RA_raac}. We defer experiments with different reward penalizations to the Appendix.

We compare EVAC over different baseline algorithms such as DDPG-RAAC \cite{RA_raac}, TD3-RAAC (TD3 version of DDPG-RAAC) and DSAC (with single critic) \cite{RA_DSAC_risk}. We freeze the architecture of the critic to be uniform across all the algorithms compared so as to allow for fair comparison. RAAC uses quantile regression to construct $Z^{\pi}(s,a)$ and uses the CVaR to optimize the policy. DSAC additionally also maximizes the entropy of the policy to ensure optimal exploration. \textbf{The difference between our approach and the compared baselines is that we additionally model the tail of the state action value distribution using the shifted and scaled GPD distribution}. We seek to find answers to the following questions: 
\begin{itemize}
    \item \textbf{Q1:} How does the GPD based tail modelling improve optimization of the risk sensitive objective?
    \item \textbf{Q2:} Does the so modelled distribution help in avoiding rare risky regions of the state action space (i.e. $(s,a)$ pairs that lead to rare risky penalizations)?
\end{itemize}

To answer \textbf{Q1}, we compute the mean CVaR (Eqn.\ref{eqn:policy_imp}) across different trained agents for each algorithm. We expect to observe higher CVaR values in agents that exhibit better risk aversion by modeling the true underlying distribution better. To answer \textbf{Q2}, we define specific metrics for each environment type (Mujoco and safety-gym) to ascertain the level of risk aversion. We define these metrics in the following subsection. Additionally, we define the cumulative reward to ascertain performance of the agent on the main task. We set the threshold quantile $\eta=0.96$ for all EVAC agents.

\input{all_tables}

\subsection{Mujoco Environments}
We primarily experiment with three Mujoco environments namely, HalfCheetah-v3, Hopper-v3 and Walker-v3 environments.
For the Half-Cheetah-v3 environment, we set the quantity $q$ as the velocity of the agent with $L=-50$, $\alpha=2.5$, $p=0.05$. For the Hopper-v3 environment, we set $q$ to denote the angle $|\theta|$ of Hopper-v3, with $L=-50$, $\alpha=0.03$ , $p=0.05$. For the Walker2d-v3 environment, we set $q$ to denote the angle $|\theta|$ of Walker2d-v3, with $L=-30$, $\alpha=0.2$ , $p=0.05$. We ensure to set the penalization rate $p$ at a low value of 5\% to simulate very rare risks.

For the Mujoco environments, we define two metrics namely, the mean overshoot and the percentage failure which are defined as :
\[\mathrm{Mean Overshoot} = \mathrm{E_N} |q - \alpha| \]
\[\mathrm{Percentage Failure} = \mathrm{E_N}[ \mathbb{I}_{q > \alpha}]*100 \]
where $\mathrm{E_N}$ represents the empirical mean over $N$ episodes, $q$ is the quantity of interest i.e. the velocity $v$ for the Half-Cheetah and the angle $|\theta|$ for the Hopper and Walker. Inference is done on 5 trained agents for the results shown in Table \ref{tab:mujoco}. Each trained agent completes an episode in inference mode acting on the learnt policy. For each step in the episode, the parameters of interest (velocity in HalfCheetah-v3) and (angles in Hopper-v3 and Walker-v3) are collected. As discussed above, we define the percentage failure to be the mean fraction of times the particular quantities of interest cross the threshold $\alpha$ across the different trained agents.  Similarly the mean overshoot is the mean overshoot amount over the defined threshold $\alpha$. .

\subsection{Safety-gym Environment}
From the safety-gym benchmark \cite{safety-gym} we use the 'Point' robot and the 'goal' task. Thus, in our setup a particular robot (the 'point' robot) is tasked with reaching a goal location on the arena. The robot receives a reward with respect to distance from the goal. However, there are certain circular regions on the arena, termed 'hazards' which may rarely lead to penalizations. The nature of reward penalization is $ \mathbf{r(s,a) = R(s,a) +} \mathbf{\mathbb{I}_{q \in \mathcal{H}} L.\mathcal{B}_p}$, where $q$ denotes the current location of the robot in the arena, $\mathcal{H}$ includes positions of all the hazards in the arena and $p$ is the Bernoulli frequency of penalization. Thus, the agent suffers a rare penalty when in the hazard region. 
We create two scenarios with respect to the placement of the hazard and the size (radius) of the hazard. 

\textbf{Safety-gym Scenario-A:} In this experimental setup, there is a single hazard of large radius placed along the straight line path (shortest distance) between the start position and the goal position. The agent needs to learn to avoid the large hazard region. We set $p=0.05$ for Scenario-A.

\textbf{Safety-gym Scenario-B:} In this more challenging experimental setup, there are multiple hazards of smaller radius placed closer to the goal. The agent needs to discover the optimal path to avoid all hazards and reach the goal. We make the setup even more challenging by setting the penalization parameter to $p=0.03$. In addition to the percentage failure, we also use the episode length as indicated in Table \ref{tab:safety-gym}.

\subsection{Result Discussion}
We express results as mean $\pm$ standard deviation across 5 trained agents. As can be seen in Table \ref{tab:mujoco}, firstly the EVAC agents exhibit conservative behavior by avoiding risky reward state actions (as is evident by the lower percentage failure and mean overshoot). 
Secondly, the CVaR of the EVAC agents is higher than the baseline methods which is why also there is higher risk aversion present in the EVAC agents. We posit that this is due to better modelling of the tail distribution by the shifted and scaled GPD distribution, which leads to precise risk averse optimization. The cumulative rewards in Table \ref{tab:mujoco} show roughly comparable performances to other baselines. In the Mujoco environments, the cumulative reward of DSAC is high (due to policy entropy) but comes at the cost of entering rare risky regions. We observe that RAAC agents are not as risk averse and do not improve cumulative performance either. EVAC agents while slightly compromising on the cumulative performance, exhibit very good risk aversion. This implies that the EVAC agents still explore systematically within the 'legal', non-risky regions to maximize cumulative reward. We observe similar trends in the safety-gym environments in Table \ref{tab:safety-gym}. The cumulative rewards collected by EVAC agents is the maximum in case of the safety-gym environments which demonstrates the EVAC agents' ability to recognize and avoid extremely rare risk hazards (at even a 3\% penalty rate). In summary, the obtained CVaR values demonstrate that EVAC agents improve optimization of the risk sensitive objective. The near zero failure percentages demonstrate that EVAC agents can also be risk averse even under extremely rare penalizations. We perform ablation studies on varying the penalization rate $p$, the threshold quantile $\eta$ etc in the Appendix. We also perform analysis and compare the estimated tail distribution of the RL agents against the ground truth distribution.  We perform further analyses and describe the experimental setup and environments in greater detail in the Appendix.

%% file: all_tables.tex
\setlength\tabcolsep{3pt}
\begin{table*}[!h]
\centering
\begin{tabular}{c c c c c c}
Environment & Algorithm & Mean Overshoot $(\downarrow)$ & Percentage Failure $(\downarrow)$ & Cumulative Reward $(\uparrow)$ & CVaR  $(\uparrow)$\\
\hline
HalfCheetah & RAAC-DDPG & 0.32$\pm$ 0.08 & 16.55$\pm$ 4.43 & 637.81$\pm$ 319.78 & 135.18 $\pm$ 13.02 \\
& RAAC-TD3 & 0.52$\pm$ 0.15 & 41.3$\pm$ 16.6 & 836.8$\pm$ 195.85 & 129.81 $\pm$ 38.75 \\
& D-SAC & 0.33 $\pm$ 0.22 & 27.43 $\pm$ 16.67 & \textbf{1522.07 $\pm$ 273.35} & 148.39 $\pm$ 10.78 \\
& \textcolor{red}{EVAC} & 0.11 $\pm$ 0.03 & \textbf{2.87 $\pm$ 1.3}  & 
\textbf{1502.46 $\pm$ 94.25} & \textbf{156.71 $\pm$ 11.07} \\
\hline
Hopper & RAAC-DDPG & 0.05$\pm$ 0.03 & 56.2$\pm$ 8.98 & 272.39$\pm$ 168.27 & 80.89 $\pm$ 79.91 \\
& RAAC-TD3 & 0.06$\pm$ 0.04 & 11.14 $\pm$ 1.46 & \textbf{307.69$\pm$ 48.49} & 49.97 $\pm$ 10.94 \\
& D-SAC & 0.05 $\pm$ 0.04 & 50.54 $\pm$ 12.40 & \textbf{373.59  $\pm$  201.35} & 90.74 $\pm$ 33.92 \\
& \textcolor{red}{EVAC} & 0.04 $\pm$ 0.02 & \textbf{3.68 $\pm$ 3.08} & 257.24 $\pm$ 34.05 & \textbf{105.66 $\pm$ 19.62} \\
\hline
Walker2d & RAAC-DDPG & 0.21$\pm$ 0.05 & 18.94$\pm$ 11.5 & 189.53$\pm$ 87.93 & 147.75 $\pm$ 64.73 \\
& RAAC-TD3 & 0.21$\pm$ 0.07 & 14.76$\pm$ 5.8 & \textbf{273.21$\pm$ 77.84} & 46.04 $\pm$ 18.68 \\
& D-SAC & 0.17 $\pm$ 0.09 &  20.18 $\pm$ 4.11 & \textbf{587.92 $\pm$ 401.4} & 123.63 $\pm$ 9.8 \\
& \textcolor{red}{EVAC} & 0.41 $\pm$ 0.1 & \textbf{2.97 $\pm$ 0.58} & 172.82 $\pm$ 16.07 &  \textbf{170.06 $\pm$ 23.55} \\
\hline
\end{tabular}
\caption{Performance metrics on Mujoco environments under a penalization rate of $p=0.05$. We record CVaR at 0.95 (0.05 if reward is not negated)}. 
\label{tab:mujoco}
\vspace{-0.25cm}
\end{table*}

\setlength\tabcolsep{1 pt}
\begin{table*}[!h]
\centering
\begin{tabular}{c c c c c c}
Environment & Algorithm & Episode Length $(\downarrow)$ & Percentage Failure $(\downarrow)$ & Cumulative Reward $(\uparrow)$ & CVaR  $(\uparrow)$\\
\hline
Safety-Gym \\Scenario-A & RAAC-DDPG  &  210.0$\pm$ 8.64 & 18.28$\pm$ 4.03 & 4.2$\pm$ 0.41 & 1.63 $\pm$ 0.06 \\
& RAAC-TD3 & 197.67$\pm$ 8.81 & 12.13$\pm$ 5.94 & 5.55$\pm$ 0.23 & 1.39 $\pm$ 0.13  \\
& D-SAC & 220.33 $\pm$ 10.5 & 10.93 $\pm$ 3.05 & 5.21 $\pm$ 0.41 & 1.62 $\pm$ 0.02 \\
& \textcolor{red}{EVAC} &  253.67 $\pm$ 17.44 & \textbf{0.0 $\pm$ 0.0} & \textbf{5.71 $\pm$ 0.0} & \textbf{1.7 $\pm$ 0.08} \\
\hline
Safety-Gym \\ Scenario-B & RAAC-DDPG &  231.0$\pm$ 19.82 &
20.2$\pm$ 14.66 & 5.22$\pm$ 0.41 & 1.58 $\pm$ 0.1 \\
& RAAC-TD3 & 216.0$\pm$ 9.42 & 22.34$\pm$ 9.93 & 4.87$\pm$ 0.47 & 0.64 $\pm$ 0.41 \\
& D-SAC &  232.33 $\pm$ 16.21 & 12.86 $\pm$ 3.37 & 5.51 $\pm$ 0.14 & 1.7 $\pm$ 0.05 \\
& \textcolor{red}{EVAC}& 223.33 $\pm$ 15.63 & \textbf{2.62 $\pm$ 3.71} & \textbf{5.61 $\pm$ 0.14} & \textbf{1.95 $\pm$ 0.12} \\
\hline

\end{tabular}
\caption{Performance metrics on the Safety-gym environment under a penalization rate of $p=0.05$ for Scenario-A and $p=0.03$ for Scenario-B. We record CVaR at 0.95 (0.05 if reward is not negated).}
\label{tab:safety-gym}
\vspace{-0.5cm}
\end{table*}

%% file: supplementary.tex
\appendix

\onecolumn

\section*{Appendix}

\section{Proof of convergence of the Bellman Operator in EVAC}

In this section, we set out to prove the convergence of the Bellman update for Equation 5 in the manuscript.

\textbf{\textit{Definition 1:}} For any two random variables $J_1(s,a)$ and $J_2(s,a)$ with distributions $Z_1(s,a)$ and $Z_2(s,a)$ with inverse CDF functions $F^{-1}_{J_1(s,a)}$ and $F^{-1}_{J_2(s,a)}$ respectively, the Wasserstein distance $d_p$ is defined as:

\[d_p\Big(F_{J_1(s,a)}, F_{J_2(s,a)} \Big) = \Big( \int_{0}^{1} |F^{-1}_{J_1(s,a)}(u) - F^{-1}_{J_2(s,a)}(u)| ^p du\Big)^{1/p}\]

Equivalently, the maximal Wasserstein distance $\Bar{d_p}$ is defined as:

\[\Bar{d_p}(F_{J_1}, F_{J_2}) = \mathrm{sup}_{s,a} d_p\Big(F_{J_1(s,a)}, F_{J_2(s,a)} \Big) \]

\textbf{\textit{Property 1:} For a scalar constant $r$, the shifted random variables 
$J_1(s,a) + r$ and $J_2(s,a) + r$  have
\[d_p\Big(F_{J_1(s,a) + r}, F_{J_2(s,a) +r} \Big) = d_p\Big(F_{J_1(s,a)}, F_{J_2(s,a)} \Big) \]
}

\textbf{ \textit{Property 2:} For a real constant scaling factor $0<\gamma<1$, the scaled random variables $\gamma J_1(s,a)$ and $\gamma J_2(s,a)$ have}
\[d_p\Big(F_{\gamma J_1(s,a)}, F_{\gamma J_2(s,a)} \Big) \leq \gamma d_p\Big(F_{J_1(s,a)}, F_{J_2(s,a)} \Big)\]

\textbf{\textit{Definition 1:}} For a distribution Z, a quantile level $\eta$ and its corresponding quantile $Z_{\eta}$, we define the non-tail distribution $Z_L = Pr(Z \leq Z_{\eta})$ and the non-tail distribution $Z_H = \frac{1}{1-\eta}Pr(Z > Z_{\eta})$.

\vspace{+0.5cm}

\textbf{ \textit{Theorem 1:}
Let $\mathcal{Z}$ denote the space of all state action value distributions. For the state action value distribution $Z(s,a) = Z_L(s,a) + (1-\eta)Z_H(s,a)$ , where $Z_L$ represents the non-tail region of $Z$ and $Z_H$ represents the tail region of $Z$ (as described in Definition 1), the Bellman operator $T^{\pi}: \mathcal{Z} \times \mathcal{Z}$, is a $\gamma$ contraction under the maximal Wasserstein distance metric $\Bar{d_p}$.}

\textbf{Note:} For notational convenience, we express $d_p\Big(F_{J_1(s,a)}, F_{J_2(s,a)} \Big)$ as $d_p\Big(Z_1(s,a), Z_2(s,a) \Big)$.

\textbf{Proof:}
\begin{align*}
    &d_p(T^{\pi}Z_1 , T^{\pi}Z_2)\\ \notag
    &=d_p \Big( r(s,a) + \gamma  \Big[ Z_{L_1}(s',a') + (1-\eta) Z_{H_1}(s',a') \Big] , r(s,a) + \gamma \Big[ Z_{L_2}(s',a') + (1-\eta) Z_{H_2}(s',a') \Big] \Big)  \\ \notag
    &=d_p \Big( \gamma  \Big[ Z_{L_1}(s',a') + (1-\eta) Z_{H_1}(s',a') \Big] , \gamma \Big[ Z_{L_2}(s',a') + (1-\eta) Z_{H_2}(s',a') \Big] \Big)  && \because \text{\textbf{Property 1}}\\ \notag
    & \leq \gamma d_p \Big(  \Big[ Z_{L_1}(s',a') + (1-\eta) Z_{H_1}(s',a') \Big] ,  \Big[ Z_{L_2}(s',a') + (1-\eta) Z_{H_2}(s',a') \Big] \Big) && \because \text{\textbf{Property 2}} \\ \notag
    &=\gamma d_p \Big( Z_1(s',a') ,  Z_2(s',a') \Big) 
    \end{align*}

\begin{align*}
    \Bar{d_p}(T^{\pi}Z_1 , T^{\pi}Z_2) &= \mathrm{sup}_{s,a}d_p \Big( T^{\pi}Z_1(s,a), T^{\pi}Z_2(s,a) \Big) \\ \notag
    & \leq \gamma \; \mathrm{sup}_{s,a}d_p \Big( Z_1(s,a), Z_2(s,a) \Big)  &&  \because \text{From Eqn. 1} \notag \\
    & = \gamma \Bar{d_p}\Big( Z_1(s,a), Z_2(s,a) \Big) \notag
\end{align*}

From the above equation, we prove that the $T^{\pi}$  operator is a contraction and that the Bellman update with the GPD tail distribution converges. 

\section{Algorithmic Representation of EVAC}

We provide the EVAC algorithm below.

\begin{algorithm}[h]
   \caption{EVAC}
   \label{alg:evt_rl}
\begin{algorithmic}
   \STATE {\bfseries Input:} Initialize $\sigma(s,a), \xi(s,a), Z(s,a)$ \\
   Select a threshold quantile level $\eta$ \\
   \textbf{POLICY ITERATION:} \\
   For tuple $(s,a,r,s, a'=\pi(s'))$ \\
    $x \sim GPD\Big(\xi(s',a'), \sigma(s',a') \Big)$ \\
    Define $Z'_H = Z(s,a)|_{\tau=\eta} + x $ \\ 
    Define $ Z'_L = Z(s,a)|_{\tau=\tau_0}; $ where $\tau_0 \sim \mathrm{Unif}(0, \eta )$\\ 
    Sample $p  \sim \mathrm{Ber}_{\eta}$ \\
    Define Bellman target: \\
    $ Z_T = r + \gamma [\mathbbm{1}_{p = 0}Z'_H + \mathbbm{1}_{p = 1}Z'_L ] $ \\
   \textit{\textbf{MLE Estimation for $\xi(s,a), \sigma(s,a)$}} \\
   $y \sim Z(s,a)|_{\tau>\eta} $ \\
   $\xi(s,a), \sigma(s,a) = \textrm{MLE}[\mathrm{GPD}(y - Z(s,a)|_{\tau=\eta}]$ \\
   \textbf{POLICY IMPROVEMENT:} \\
   Update policy $\pi$ according to Eqn.7 (in manuscript)
   
\end{algorithmic}
\end{algorithm}

\section{MLE Estimation of the parameters of the GPD Distribution}

The CDF of the GPD distribution $F_{\xi, \sigma}(x)$ is given by:
\begin{equation*}
    \begin{cases}
  1- \Big( 1+ \frac{\xi x}{\sigma} \Big)^{-1/\xi} \; \mathrm{for} \; \xi \neq 0\\    
  1- \exp(-\frac{x}{\sigma}) \; \mathrm{for} \; \xi =0    
\end{cases}
\end{equation*}

The log-density function (log-PDF) of the same GPD distribution is given by:

\begin{equation}
\log f_{\xi,\sigma}(x) = 
    \begin{cases}
  -\log (\sigma) + \Big(-1/\xi - 1 \Big) \log \Big(1 + \xi x/\sigma \Big) \;\mathrm{for}  \; \xi \neq 0\\    
  -\log (\sigma) - x/\sigma \; \mathrm{for} \; \xi =0    
\end{cases}
\end{equation}

\begin{equation*}
    \frac{\partial \log f_{\xi, \sigma}(x)}{\partial \xi}  = \Big( -1/\xi -1\Big) \Big(\frac{1}{1 + \xi x/\sigma} \Big) \cdot \frac{x}{\sigma} + \log\Big( 1 + \xi x/\sigma \Big) \Big( 1/ \xi^2\Big) = 0
\end{equation*}

\begin{equation*}
    \frac{\partial \log f_{\xi, \sigma}(x)}{\partial \sigma}  = -\frac{1}{\sigma} + \Big( -1/\xi -1\Big) \cdot \frac{1}{1+\xi x/\sigma} \; \cdot \xi x
 =0
 \end{equation*}

 However when $\xi=0$; we are left with the MLE estimation of the parameter $\sigma$ of the exponential distribution

 \begin{equation}
     \frac{\partial \log f_{0, \sigma}(x)}{\partial \sigma} = -\frac{1}{\sigma} + \frac{x}{\sigma^2} = 0
 \end{equation}

We make the GPD parameter $\sigma(s,a), \xi(s,a)$ be parameterized and represent it by $\sigma_{\theta}(s,a), \xi_{\phi}(s,a)$.
Given a batch of transition tuples $(s_i,a_i,r_i,s'_i) \; ; i=1 \to B$, where $B$ is the batch size, we source samples from the GPD distribution by sampling $K$ samples $x_k$ from  
$Z^{\pi}_H(s,a)$, the constructed GPD distribution. We then compute the empirical log-likelihood loss $\mathcal{L}$ that needs to be maximized.

\begin{equation}
    \mathcal{L}_{\phi} =  \frac{1}{B} \sum_{i=1}^{B} \cdot  \frac{1}{K}\sum_{k=1}^{K} \Big[ \frac{\partial \log f_{\xi, \sigma}(x_k)}{\partial \xi_{\phi}} \Big]
\end{equation}

\begin{equation}
    \mathcal{L}_{\theta} =  \frac{1}{B} \sum_{i=1}^{B} \cdot  \frac{1}{K}\sum_{k=1}^{K} \Big[ \frac{\partial \log f_{\xi, \sigma}(x_k)}{\partial \sigma_{\theta}} \Big]
\end{equation}

where $x_k \sim Z^{\pi}_H(s,a)$, $K$ is the number of samples sampled from the GPD distribution $Z^{\pi}_H(s,a)$. We set $K=100$ and $B=128$ for all experimentation. We perform gradient ascent on the parameters $\theta$ by using the empirical loss:
\begin{align}
    &\theta : \theta + \alpha_{lr} \cdot  \mathcal{L}_{\theta}\\ \notag
    &\phi : \phi + \alpha_{lr} \cdot \mathcal{L}_{\phi}
\end{align}

where $\alpha_{lr}$ is the learning rate.

\section{Experimental details and Hyperparameter setting}

We use an actor critic framework , where the critic is a quantile critic. Both the actor and critic have 3 layers with hidden size being 128. 

\subsection{Mujoco Environments}
During training and inference, the max episode length of the agent is set to 1000. During training, the agents were trained for 100,000 time steps on the whole. The batch size $B=128$ and we set $K$, the number of samples sampled from the GPD distribution to 50. We set the learning rates for the actor and critic to $0.001$ in all cases. The discount factor $\gamma=0.99$ for all cases too. The soft update parameter $\tau=0.02$ for all our experiments on the Hopper-v3 and Walker2d-v3, while $\tau=0.01$ for the HalfCheetah-v3 environment.

\subsection{Safety Gym Environments}
During training and inference, the max episode length of the agent is set to 1000. During training, the agents were trained for 70,000 time steps on the whole for the safety-gym suite.

In all of the ablation studies, we report our results on 3 different trained agents. 

\section{Ablation Study: Risk Awareness and Performance with changing Penalization rate $p$}

We recall that the penalization rate $p$ is defined as the Bernoulli distribution parameter that is used to create rare risky rewards of the form $r(s,a) = R(s,a) + \mathbf{1}_{q > \alpha} L.\mathcal{B}_p$, as defined in the manuscript. As $p$ becomes smaller, the penalty becomes rarer. 

In this section we change this penalization rate $p$ of the reward for the HalfCheetah-v3 environment 
$r(s,a) = R(s,a) + \mathbf{1}_{v > \alpha} L.\mathcal{B}_p$. We move from a very rare reward penalty of $p=0.03$ to a fairly frequent penalty rate of $p=0.1$. For our experimentation purposes, we fix the threshold quantile $\eta=0.96$ for both EVAC and RAAC-TD3. \textbf {From Table \ref{tab:varying_p} however, as the penalization rate $p$ becomes smaller (rare risky events), the EVAC agent still exhibits very small percentages of failure while the cumulative reward is still very high. This bolsters the fact that the EVAC agent still explores reasonably while being risk averse.}   

\begin{table*}[!h]
\centering
\begin{tabular}{ c c c c c c}
 Algorithm &$p$ & Mean Overshoot $(\downarrow)$ & Percentage Failure $(\downarrow)$ & Cumulative Reward $(\uparrow)$ & CVaR  $(\uparrow)$\\
\hline
 EVAC &0.03 & 0.09 $\pm$ 0.05 & \textbf{6.43 $\pm$ 4.99} & 1412.81 $\pm$ 292.22 & 140.66 $\pm$ 23.57\\
& 0.05  & 0.11 $\pm$ 0.03 &\textbf{2.87 $\pm$ 1.3}  & 
1502.46 $\pm$ 94.25 & 156.71 $\pm$ 11.07\\
&0.1   & 0.09 $\pm$ 0.0 & \textbf{1.07 $\pm$ 0.63} & 1160.8 $\pm$ 264.46 & 111.48 $\pm$ 33.65 \\
\hline
 RAAC-TD3 & 0.03 & 0.85 $\pm$ 0.39 & 30.23 $\pm$ 12.21 & 847.86 $\pm$ 145.21 & 85.66 $\pm$ 24.12\\
& 0.05 & 0.52$\pm$ 0.15 & 41.3$\pm$ 16.6 & 836.8$\pm$ 195.85 & 129.81 $\pm$ 38.75 \\
& 0.1 & 0.32 $\pm$ 0.11 & 13.97 $\pm$ 5.58 & 366.0 $\pm$ 453.85 & 104.92 $\pm$ 7.08 \\
\hline
\end{tabular}
\caption{Table showing various performance metrics as the penalization
rate $p$ is varied for the HalfCheetah-v3 environment (with fixed threshold quantile $\eta = 0.96$).}. 
\label{tab:varying_p}
\vspace{-0.25cm}
\end{table*}

\section{Comparison of the modelled distribution with the Ground-truth distribution}

Since we believe that the use of EVT theory helps in better modelling the tail of the distribution of the future sum of discounted returns, we compare the modelled distribution $Z^{\pi}(s,a)$ of every algorithm against the true empirically obtained distribution of the sum of future discounted returns.

We freeze the initial state to be the same across all algorithms (RAAC-DDPG, RAAC-TD3, D-SAC, EVAC) and collect trajectories according to optimal policy defined for each algorithm. For example, each algorithm has a defined optimal policy and a trajectory $\mathcal{T}^t= \{R_1, R_2, \cdots\}$ is collected by following the optimal policy of that algorithm. For such a trajectory $\mathcal{T}^t$, we compute the future discounted sum of rewards $J^t=\sum_i \gamma^iR_i$ for the very first state in the trajectory. We repeat this process of trajectory collection for the same initial state $N$ times to estimate the distribution of the sum of future discounted returns for each algorithm from $\{J^t\}_{t=1}^{N}$.

On the other hand we sample values from the quantile critic of each algorithm to obtain the modelled distribution $Z^{\pi}(s,a)$. We collect $\{z^j\}_{j=1}^{M}$, where $z^j \sim Z(s,\pi(s))$. It is to be noted that since each algorithm has a different optimal policy, the empirical distribution of the sum of future discounted returns would vary for each algorithm. 

\begin{figure}[!h]
    \centering
    
    \begin{subfigure}{0.6\textwidth}
        \includegraphics[width=\linewidth , height=4cm]{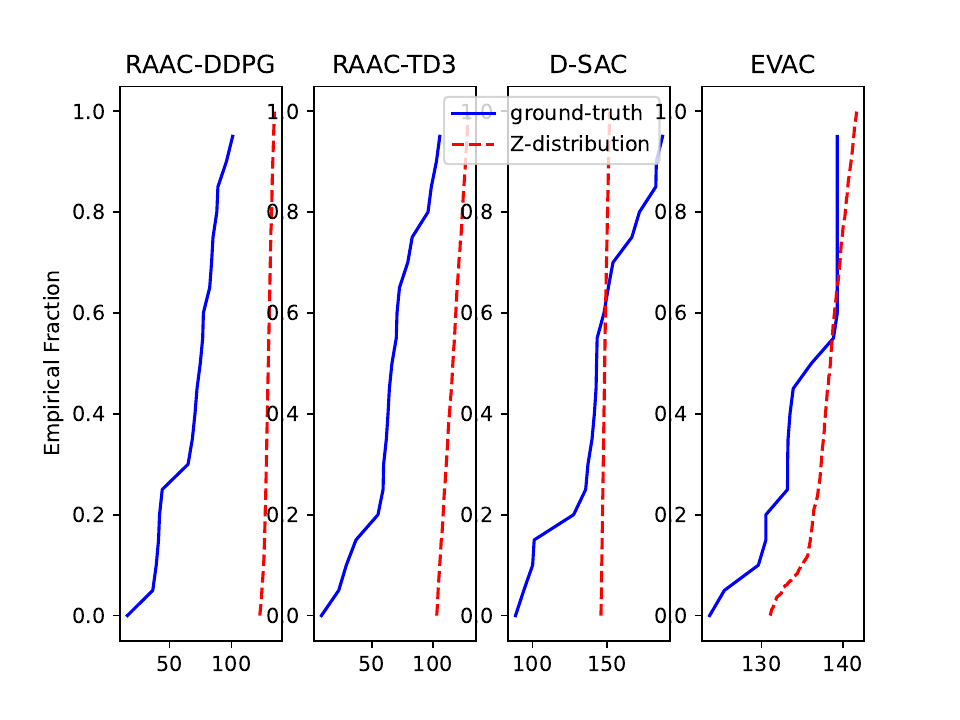}
        \caption{HalfCheetah-v3}
        \label{subfig:dist_cheetah}
    \end{subfigure}
    
    
    \begin{subfigure}{0.6\textwidth}
        \includegraphics[width=\linewidth, height=4cm]{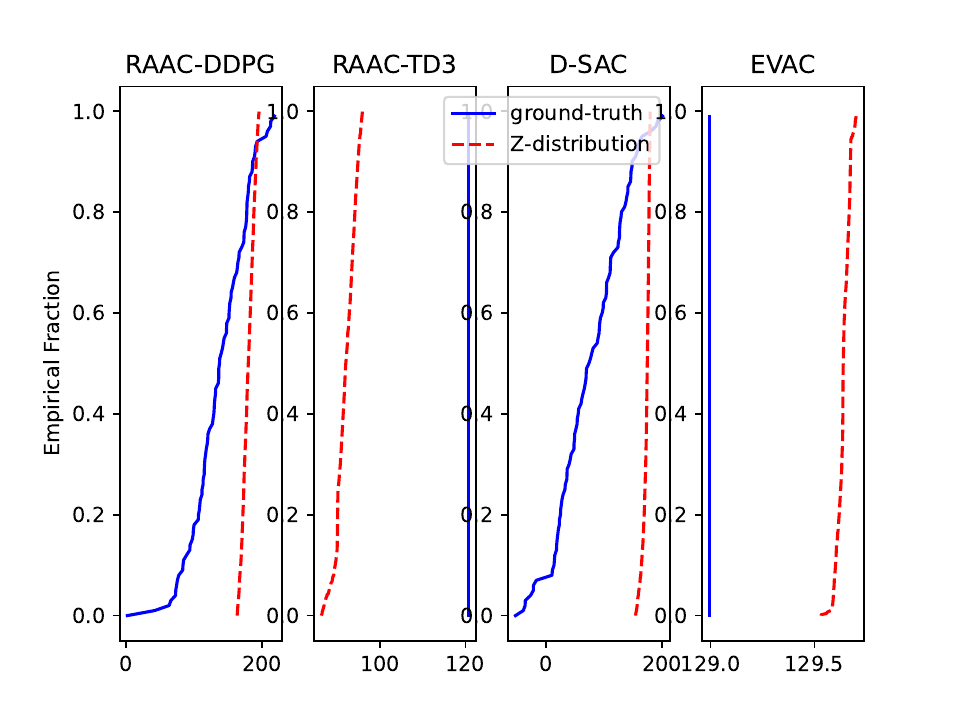}
        \caption{Hopper-v3}
        \label{subfig:dist_hopper}
    \end{subfigure}
    
    
    \begin{subfigure}{0.6\textwidth}
        \includegraphics[width=\linewidth, height=4cm]{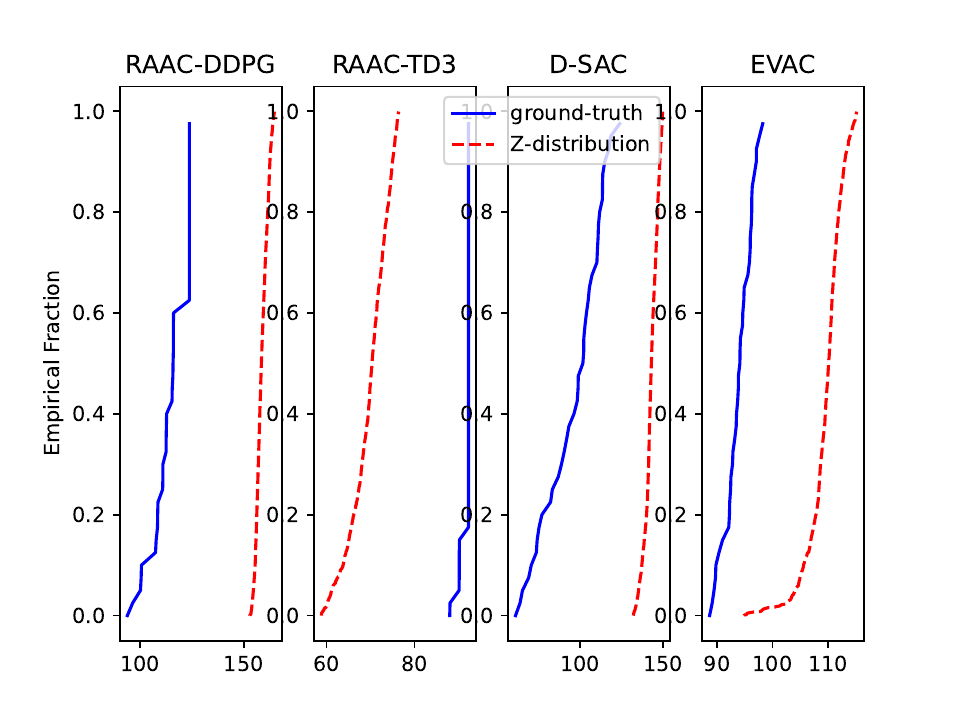}
        \caption{Walker2d-v3}
        \label{subfig:dist_walker}
    \end{subfigure}
    
    \caption{Comparison of the empirical CDF of the modelled distribution (in dashed red) and the empirical CDF of the ground truth (in solid blue) sum of discounted rewards. All plots are plotted for non-negated rewards. The tail behavior (left tail) of the EVAC agents almost precisely matches the true empirical distribution of the sum of the future discounted rewards. (Note that the scale of the X-axis is different for each algorithm and environment) }
    \label{fig:dist}
\end{figure}

We then plot the empirical CDF of both $\{J^t\}_{t=1}^{N}$ (which is called the ground truth) and $\{z^j\}_{j=1}^{M}$ (modelled Z-distribution). These empirical CDFs are plotted for each algorithm and for each Mujoco agent in \textbf{Figure \ref{fig:dist}}. 

\begin{table}[!htbp]
\setlength{\tabcolsep}{0.5\tabcolsep}
    \centering
    \begin{tabular}{c c c}
         Environment & Algorithm & 1-Wasserstein distance ($\downarrow$)\\
          \hline
          HalfCheetah-v3 & RAAC-DDPG & 55.05 \\
           & RAAC-TD3 &  57.69\\
           & D-SAC & 37.55\\
           & EVAC &  \textbf{14.89}\\
          \hline
          Hopper-v3 & RAAC-DDPG & 38.22 \\
           & RAAC-TD3 &  28.77\\
           & D-SAC & 86.87 \\
           & EVAC & \textbf{0.64} \\
          \hline
          Walker2d-v3 & RAAC-DDPG  & 42.70\\
           & RAAC-TD3 & 21.95\\
           & D-SAC & 47.40\\
           & EVAC & \textbf{3.32} \\
          \hline
    \end{tabular}
    \caption{Table showing 1-Wasserstein distance between the empirical ground truth distribution of the sum of discounted rewards and the modelled distribution as shown in Figure \ref{fig:dist} }
    \label{tab:wasserstein_dist}
\end{table}

We compute the 1-Wasserstein distance between the ground truth and the modelled distribution in Table \ref{tab:wasserstein_dist}. The 1-Wasserstein distance $d_1(X,Y)$ between two random distributions $X$ and $Y$ is defined as:

\[d_1(X, Y) = \Big( \int_{0}^{1} |F^{-1}_{X}(u) - F^{-1}_{Y}(u)| du\Big)\]

Firstly for the HalfCheetah-v3, we see that the left tail of the ground truth and the modelled distribution for EVAC, both concentrate in the same region of around 100-120. For the Hopper-v3, the ground truth and the modelled Z distribution both are almost perfectly aligned at a value of around 129 -130. For the Walker2d-v3, we see a similar trend in the tails of the groundtruth and modelled distribution for the EVAC algorithm. (Note that the scale of the X-axis is different for each algorithm and environment). 
\textbf{ From Table \ref{tab:wasserstein_dist} and Figure \ref{fig:dist}, we notice that there is generally a smaller discrepancy between the ground-truth distribution of the EVAC agent and its corresponding  modelled distribution, in comparison with other baselines that do not use EVT to model the tail behavior. Thus, EVAC agents are able to better model the true underlying distribution.}

\section{Ablation Study: Risk Awareness and Performance with changing threshold quantile parameter $\eta$}

We next address the sensitivity to risk awareness when the threshold quantile parameter $\eta$ is changed for the HalfCheetah-v3 environment. The value of $\eta$ indicates the threshold quantile level after which the GPD approximation for $Z^{\pi}_H(s,a)$ is made. 

We fix the penalization rate $p=0.05$ while changing $\eta$ continuously. We also fix $x_1=0.95$ and $x_2=1.0$ for CVaR calculation (see next section).

\begin{table*}[!h]
\centering
\begin{tabular}{c c c c c c}
Environment & $\eta$ & Mean Overshoot $(\downarrow)$ & Percentage Failure $(\downarrow)$ & Cumulative Reward $(\uparrow)$ & CVaR  $(\uparrow)$\\
\hline
HalfCheetah-v3 & 0.90 & 0.2 $\pm$ 0.04 & 35.17 $\pm$ 23.26 & 946.65 $\pm$ 234.75 & 183.68 $\pm$ 35.38\\
& 0.96  & 0.11 $\pm$ 0.03 & \textbf{2.87 $\pm$ 1.3 } & 
1502.46 $\pm$ 94.25 & 156.71 $\pm$ 11.07\\
\hline

\end{tabular}
\caption{Table illustrating various performance metrics as the threshold quantile $\eta$ is varied for the HalfCheetah-v3 environment (with fixed penalization rate $p = 0.05$).}
\label{tab:varying_eta}
\vspace{-0.25cm}
\end{table*}

We observe that the choice of the threshold quantile $\eta$ plays an important role in risk mitigation too. We observe that risk sensitivity increases with increase in the value of $\eta$. We posit that this is due to better approximation of the GPD distribution as $\eta$ increases (Refer Theorem \ref{th:PBD} in the manuscript). \textbf{From Table \ref{tab:varying_eta} as $\eta$ increases, the EVAC agents become very risk averse while producing good cumulative rewards. }. This underscores the effectiveness of using EVT theory for risk mitigation in reinforcement learning. 

\section{Effect of increasing $x_1$ in CVaR optimization (more conservative behavior)}

We recall from Equation 7 in the manuscript that $x_1$ is used to control the level of risk aversion. We re-write equation 7 from the manuscript below.

\begin{align*}
\label{eqn:policy_imp}
    \pi^{*} &=  \argmin_{\pi} \mathrm{CVaR(x_1)} \\ \notag
    &=\argmin_{\pi} \frac{1}{x_2 - x_1}\int_{\tau=x_1}^{x_2=1} Z\Big(s,\pi(s)\Big) \Big|_{\tau}
\end{align*}

As $x_1 \to 1$, the policy is incentivized to optimize the right tail and is thus made more risk averse. Instead of $x_1=0.95$, we set $x_1=0.99$ to ensure the highest level of conservatism.

\begin{table*}[!h]
\centering
\begin{tabular}{c c c c c c}
Environment & Algorithm & Mean Overshoot $(\downarrow)$ & Percentage Failure $(\downarrow)$ & Cumulative Reward $(\uparrow)$ & CVaR  $(\uparrow)$\\
\hline
HalfCheetah-v3 & RAAC-DDPG & 0.97 $\pm$ 0.68 & 47.07 $\pm$ 17.64 & 677.15 $\pm$ 173.03 & 126.22 $\pm$ 115.43\\
& RAAC-TD3   & 0.58 $\pm$ 0.17 & 30.47 $\pm$ 4.97 & 360.55 $\pm$ 242.4 & 104.21 $\pm$ 12.0 \\
& D-SAC  & 0.25 $\pm$ 0.19 & 35.7 $\pm$ 40.42 & 1089.13 $\pm$ 395.36 & 143.37 $\pm$ 50.41 \\
& EVAC  & 0.03 $\pm$ 0.02 & \textbf{1.03 $\pm$ 1.26} & 872.62 $\pm$ 171.62 & 151.45 $\pm$ 4.33\\
\hline

\end{tabular}
\caption{Table showing the the percentage failure and also the cumulative reward as risk sensitivity parameter $x_1$ is  made stricter at 0.99 instead of 0.95 for the HalfCheetah-v3 environment (with fixed penalization rate $p = 0.05$ and threshold quantile $\eta$=0.96).}
\label{tab:varying_x1}
\vspace{-0.25cm}
\end{table*}

From Table \ref{tab:varying_x1}, even when the CVaR objective is made very risk averse (at $x_1=0.99$), the baseline agents still do not demonstrate good risk mitigation. The vanilla quantile based methods which depend on sampling exclusively, may not model the tail behavior accurately enough. This implies that even if $x_1$ is made arbitrarily large (more risk averse) in such agents, proper risk mitigation may still not be observed. EVAC on the other hand demonstrates good risk aversion. This further acts as a motivation to using EVAC-like agents which seem to evidently show better risk mitigation.

\section{ Experiments with non-Bernoulli penalizations}

In the manuscript, we model rare risky events using a Bernoulli penalty. However, there could be other risk modelling strategies. To test EVAC's ability to generalize to different risk-modelling choices, we experiment with a different penalty term in this section. We experiment with two environments namely HalfCheetah-v3 and Hopper-v3. We define the new risky reward as below.

\textbf{HalfCheetah-v3}: 
\[r(s,a) = R(s,a) - 2*\mathbf{1}_{v > \alpha} \mathrm{abs} \Big( \mathcal{N}(0, 0.5) \Big)\]

where $\alpha=2.5$.

\textbf{Hopper-v3}: 
\[r(s,a) = R(s,a) - \mathbf{1}_{|\theta| > \alpha} \mathrm{abs} \Big(\mathcal{N}(0, 0.5) \Big)\]

where $\alpha=0.03$.

\begin{table*}[!h]
\centering
\begin{tabular}{c c c c c c}
Environment & Algorithm & Mean Overshoot $(\downarrow)$ & Percentage Failure $(\downarrow)$ & Cumulative Reward $(\uparrow)$ & CVaR  $(\uparrow)$\\
\hline
HalfCheetah-v3 & RAAC-DDPG & 0.18 $\pm$ 0.03 & 1.7 $\pm$ 0.86 & 1657.61 $\pm$ 71.14 & 165.95 $\pm$ 11.69 \\
& RAAC-TD3  & 0.7 $\pm$ 0.38 & 48.57 $\pm$ 28.08 & 1791.17 $\pm$ 589.42 & 124.78 $\pm$ 20.24\\
& D-SAC  & 0.1 $\pm$ 0.06 & 12.2 $\pm$ 5.03 & 1912.31 $\pm$ 50.78 & 209.98 $\pm$ 3.66\\
& EVAC  & 0.04 $\pm$ 0.03 & \textbf{0.87 $\pm$ 0.94} & 1300.59 $\pm$ 167.26 & 116.07 $\pm$ 12.78 \\
\hline
Hopper-v3 & RAAC-DDPG&  0.03 $\pm$ 0.01 & 27.68 $\pm$ 8.26 & 1190.97 $\pm$ 323.52 & 3.87 $\pm$ 1.72\\
& RAAC-TD3  & 0.05 $\pm$ 0.01 & 56.85 $\pm$ 18.4 & 815.38 $\pm$ 503.66 & 134.72 $\pm$ 99.82\\
& D-SAC  & 0.02 $\pm$ 0.02 & 17.36 $\pm$ 12.87 & 1925.71 $\pm$ 852.54 & 114.53 $\pm$ 145.44\\
& EVAC  & 0.03 $\pm$ 0.01 & \textbf{2.29 $\pm$ 2.1} & 1820.06 $\pm$ 1075.15 & 117.65 $\pm$ 155.82 \\
\hline
\end{tabular}
\caption{Performance metrics on Mujoco environments under a penalty that is distributed as an absolute Gaussian. We record CVaR at 0.95 (0.05 if reward is not negated)}
\label{tab:risk_averse_abs_Gauss}
\end{table*}

The absolute valued Gaussian samples simulate instances where very risky rewards have a finite probability of being sampled. Unlike the Bernoulli penalty, the penalty distribution is continous valued. As can be seen, the EVAC agent is able to show better risk mitigation in comparison to all the baseline methods.

\section{Trajectory Visualization on Safety-gym environments }

\begin{figure}[!h]
    \centering
    
    \begin{subfigure}{0.45\textwidth}
        \includegraphics[width=\linewidth,height=4cm]{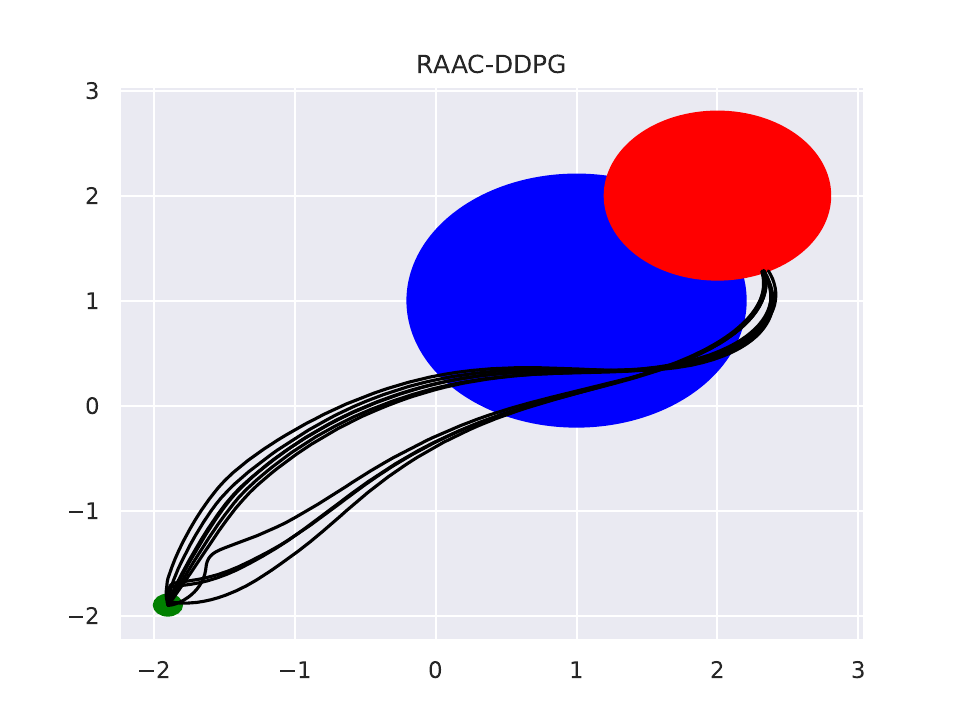}
        \caption{RAAC-DDPG}
        \label{subfig:fig1}
    \end{subfigure}
    \begin{subfigure}{0.45\textwidth}
        \includegraphics[width=\linewidth,height=4cm]{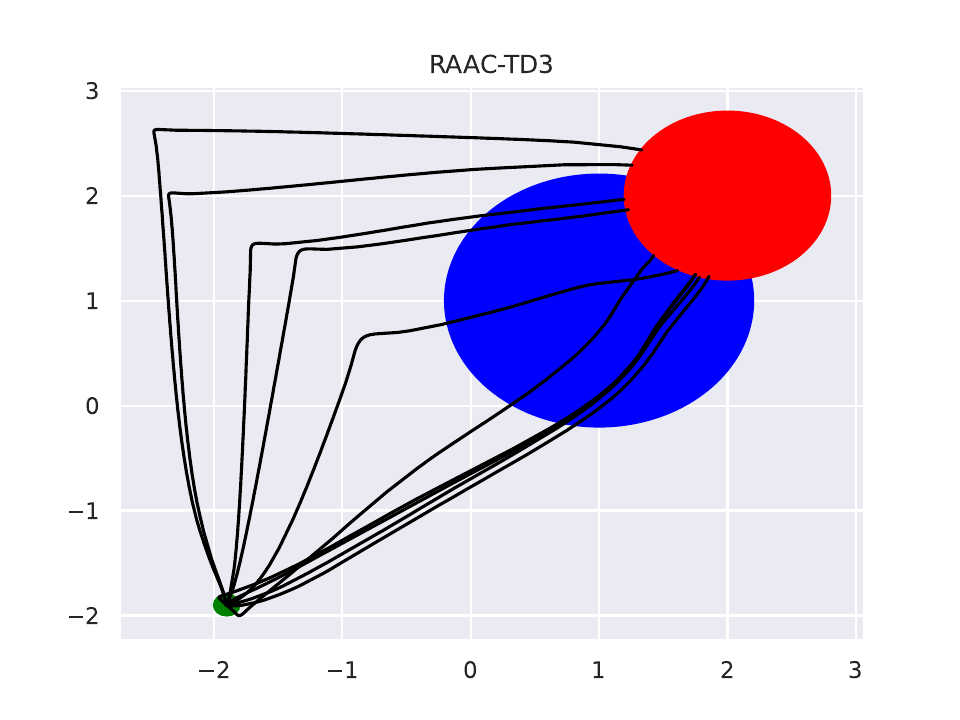}
        \caption{RAAC-TD3}
        \label{subfig:fig2}
    \end{subfigure}
    
    
    \begin{subfigure}{0.45\textwidth}
        \includegraphics[width=\linewidth,height=4cm]{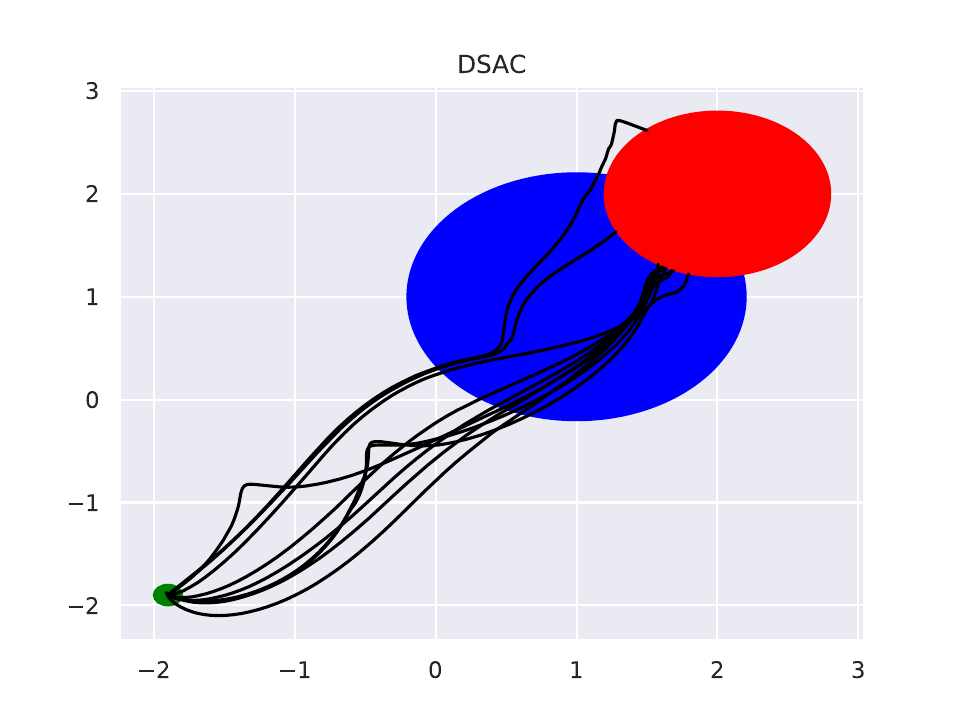}
        \caption{DSAC}
        \label{subfig:fig3}
    \end{subfigure}
    \begin{subfigure}{0.45\textwidth}
        \includegraphics[width=\linewidth,height=4cm]{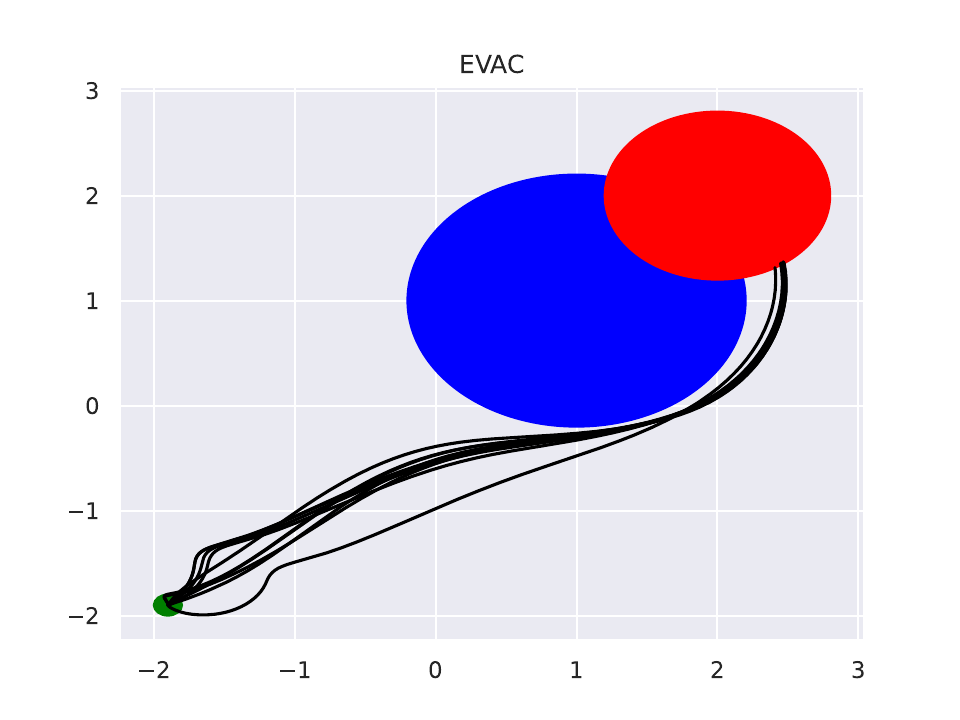}
        \caption{EVAC}
        \label{subfig:fig4}
    \end{subfigure}
    
    \caption{Trajectories visualized for the Safety-gym Scenario -A with $p=0.05$. The EVAC agent successfully navigates to the goal by recognizing the blue hazards.}
    \label{fig:scenario-A}
\end{figure}

\begin{figure}[!h]
    \centering
    
    \begin{subfigure}{0.45\textwidth}
        \includegraphics[width=\linewidth,height=4cm]{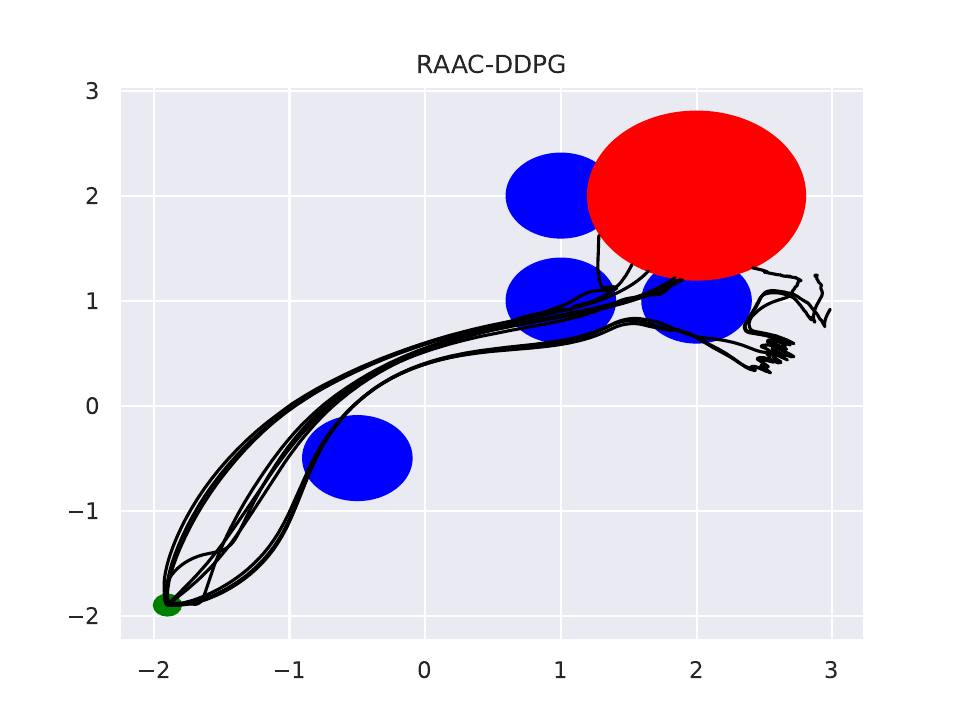}
        \caption{RAAC-DDPG}
        \label{subfig:fig1}
    \end{subfigure}
    \begin{subfigure}{0.45\textwidth}
        \includegraphics[width=\linewidth,height=4cm]{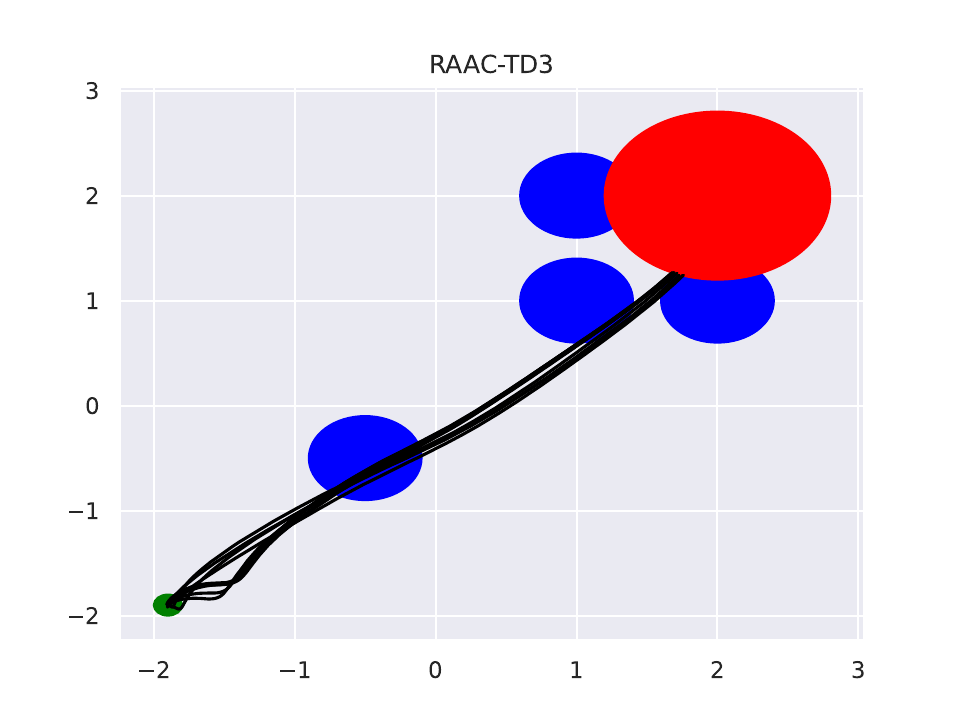}
        \caption{RAAC-TD3}
        \label{subfig:fig2}
    \end{subfigure}
    
    
    \begin{subfigure}{0.45\textwidth}
        \includegraphics[width=\linewidth,height=4cm]{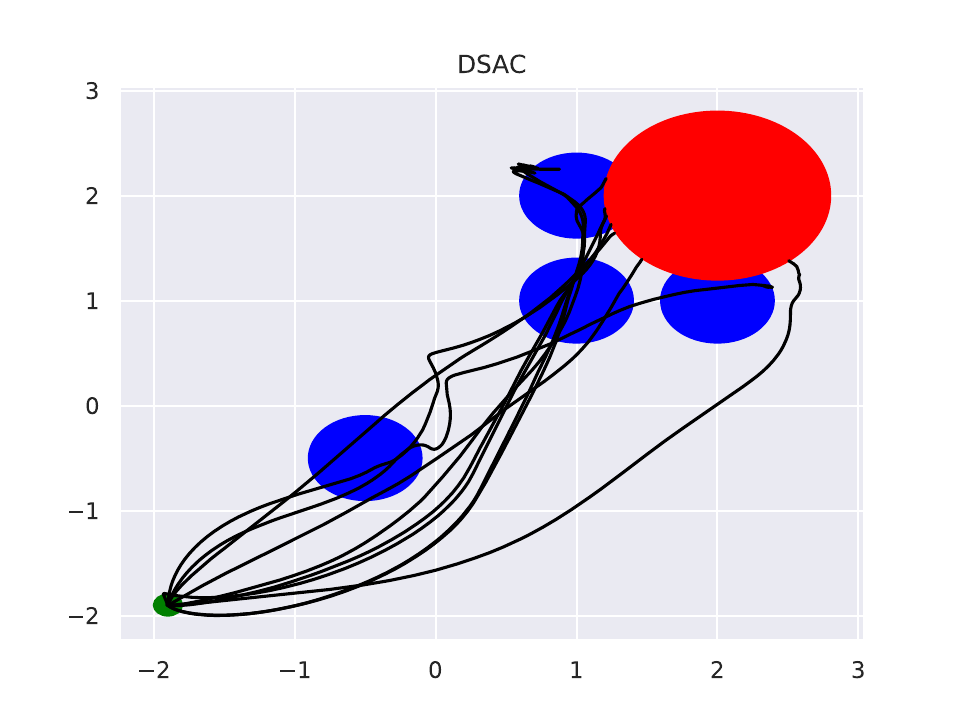}
        \caption{DSAC}
        \label{subfig:fig3}
    \end{subfigure}
    \begin{subfigure}{0.45\textwidth}
        \includegraphics[width=\linewidth,height=4cm]{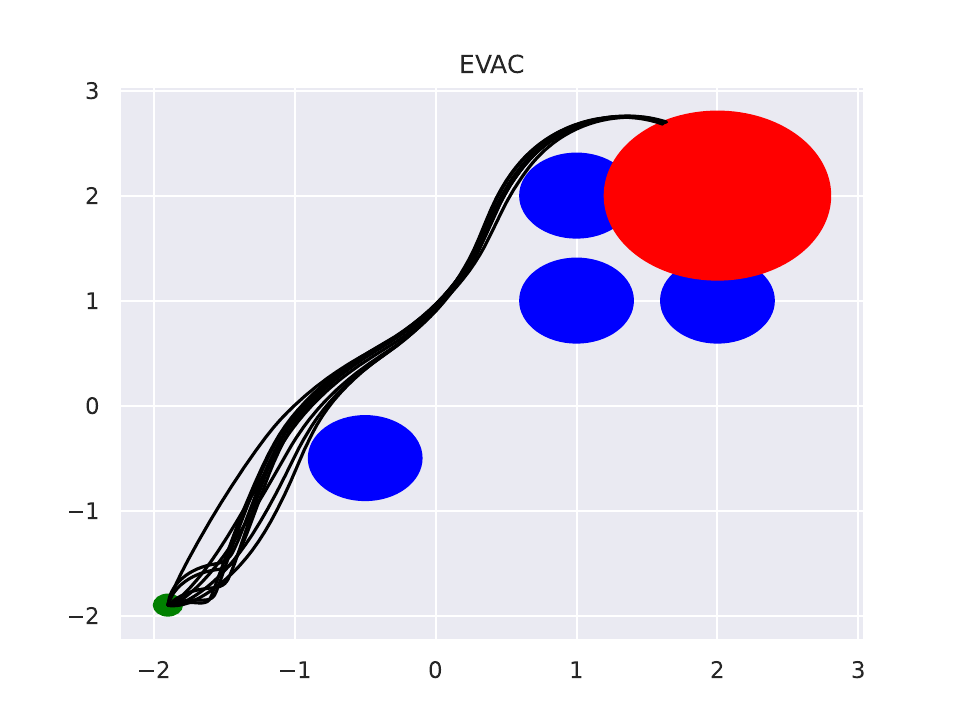}
        \caption{EVAC}
        \label{subfig:fig4}
    \end{subfigure}
    
    \caption{Trajectories visualized for the Safety-gym Scenario-B with $p=0.03$. The EVAC agent successfully navigates to the goal by recognizing the blue hazards.}
    \label{fig:Scenario-B}
\end{figure}

We represent the hazards by blue circles. The starting position is indicated in green and the goal position is indicated in red in Figures \ref{fig:scenario-A} and \ref{fig:Scenario-B}. As discussed in the manuscript, the goal of the agent is to navigate to the goal while avoiding the blue 'hazard circles'. In the two scenarios A and B, we observe that the EVAC agent is able to efficiently learn the locations of the hazard and avoid such hazards at all times. However, due to the extremely rare penalty rate of $p=0.05$ for Scenario-A and $p=0.03$ for Scenario-B, the other baseline agents, do not seem to recognize the low probability events and often enter the hazardous blue circles.

\section{Limitations and Future Work}
The use of EVT in for risk mitigation in distributional reinforcement is novel. Our work tries to illustrate the benefit of using EVT theory in risk aware reinforcement learning. One of the challenges in the use of EVT RL is the overhead involved in training parameters of the GPD distribution through maximum likelihood estimation. The added overhead may prove expensive for both training and inference, especially on edge devices. We plan to further investigate the usefulness of EVT RL in other domains and study its scaling in other complex environments.